\documentclass{article}

\usepackage{microtype}
\usepackage{graphicx}
\usepackage{subcaption}
\usepackage{booktabs} 

\usepackage{hyperref}

\usepackage[accepted]{icml2026}

\usepackage{amsmath}
\usepackage{amssymb}
\usepackage{mathtools}
\usepackage{amsthm}

\usepackage[capitalize,noabbrev]{cleveref}

\theoremstyle{plain}

\theoremstyle{definition}

\theoremstyle{remark}

\usepackage{microtype}
\usepackage{graphicx}
\usepackage{booktabs} 

\usepackage{booktabs}
\usepackage{multirow}

\usepackage{xspace}
\newcommand{\ours}{UPipe\xspace}

\usepackage{graphicx}
\usepackage{float}
\usepackage{placeins} 

\usepackage{multirow}
\usepackage{array}
\usepackage{booktabs}

\usepackage{tikz}
\usepackage{xcolor} 

\newcommand{\ballnumber}[1]{%
  \tikz[baseline=(myanchor.base)]
    \node[circle, fill=black, inner sep=1pt] (myanchor)
      {\color{white}\bfseries\footnotesize #1};%
}

\usepackage{xcolor}

\usepackage{graphicx}
\usepackage{subcaption}  
\usepackage{multirow}    
\usepackage{array}       

\icmltitlerunning{Untied Ulysses: Memory-Efficient Context Parallelism via Headwise Chunking}

\begin{document}

\twocolumn[
  \icmltitle{Untied Ulysses: Memory-Efficient Context Parallelism via Headwise Chunking}



  \icmlsetsymbol{equal}{*}

  \begin{icmlauthorlist}
    \icmlauthor{Ravi Ghadia}{together}
    \icmlauthor{Maksim Abraham}{together}
    \icmlauthor{Sergei Vorobyov}{together}
    \icmlauthor{Max Ryabinin}{together}
  \end{icmlauthorlist}

  \icmlaffiliation{together}{Together AI}

  \icmlcorrespondingauthor{Ravi Ghadia}{rghadia@utexas.edu}
  \icmlcorrespondingauthor{Max Ryabinin}{mryab@together.ai}

  \icmlkeywords{Context Parallelism, Long-Sequence Training, Transformer models}

  \vskip 0.3in
]



\printAffiliationsAndNotice{}  

\begin{abstract}
Efficiently processing long sequences with Transformer models usually requires splitting the computations across accelerators via context parallelism.
The dominant approaches in this family of methods, such as Ring Attention or DeepSpeed Ulysses, enable scaling over the context dimension but do not focus on memory efficiency, which limits the sequence lengths they can support.
More advanced techniques, such as Fully Pipelined Distributed Transformer or activation offloading, can further extend the possible context length at the cost of training throughput.
In this paper, we present \textbf{\ours{}}, a simple yet effective context parallelism technique that performs fine-grained chunking at the attention head level.
This technique significantly reduces the activation memory usage of self-attention, breaking the activation memory barrier and unlocking much longer context lengths.
Our approach reduces intermediate tensor memory usage in the attention layer by as much as \textbf{87.5$\%$} for 32B Transformers, while matching previous context parallelism techniques in training speed.
\ours{} can support the context length of \textbf{5M} tokens when training Llama3-8B on a single 8$\times$H100 node, improving upon prior methods by over \textbf{25$\%$}.
\end{abstract}

\section{Introduction}

The Transformer architecture~\cite{transformer} has powered significant advances in AI in recent years, ranging from language models with agentic and reasoning capabilities~\cite{gemini3,kimik2,minimaxm1} to video generation~\cite{wan,hunyuanvideo,magi1}.
As the field continues to progress, the demand for longer context lengths in AI models continues to grow due to applications such as code generation~\cite{starcoder, qwencoder}, long document understanding~\cite{mlongdoc, longrag}, or even audio processing~\cite{audio}.
However, training models to effectively process such long sequences is limited by the accelerator hardware: beyond a certain limit, even keeping the activations necessary for self-attention becomes a bottleneck.
As a result, methods that reduce the memory requirements of long-context training have recently attracted a surge of research interest.

The most scalable approaches for increasing the context size beyond a single accelerator leverage distributed training, splitting the computations and memory allocations across multiple devices.
In particular, the context parallelism~\cite{seqparallel, ring, ulysses} family of methods (also known as sequence parallelism) focuses on sharding model operations across the sequence axis.
These methods enable effective scaling in context length with the number of accelerators, but the activation memory per device still scales linearly with the sequence length.
Therefore, at very long sequence lengths ($>$2M), the activation memory starts to become a bottleneck, limiting the training capacity.

\begin{figure}[t]
    \centering
    \includegraphics[width=\linewidth]{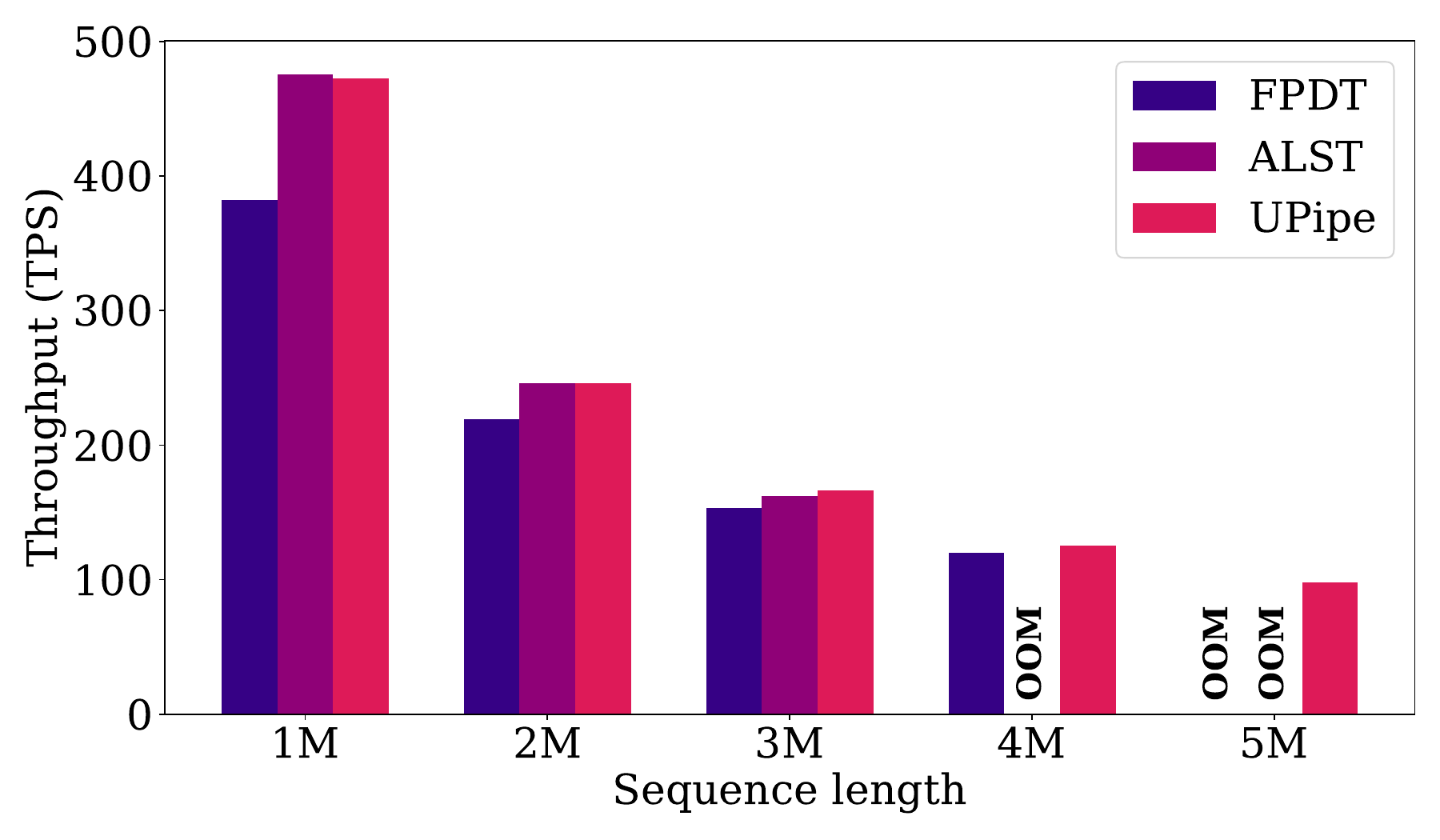}
    \caption{Comparison of context parallelism approaches on long-sequence training for Llama3-8B using 8 $\times$ H100s. UPipe provides maximum efficiency, resulting in a longer maximum context length (5M tokens) while retaining throughput.}
    \label{fig:placeholder}
    \vspace{-1em}
\end{figure}

\begin{figure*}[t]
    \centering
        \includegraphics[width=\linewidth]{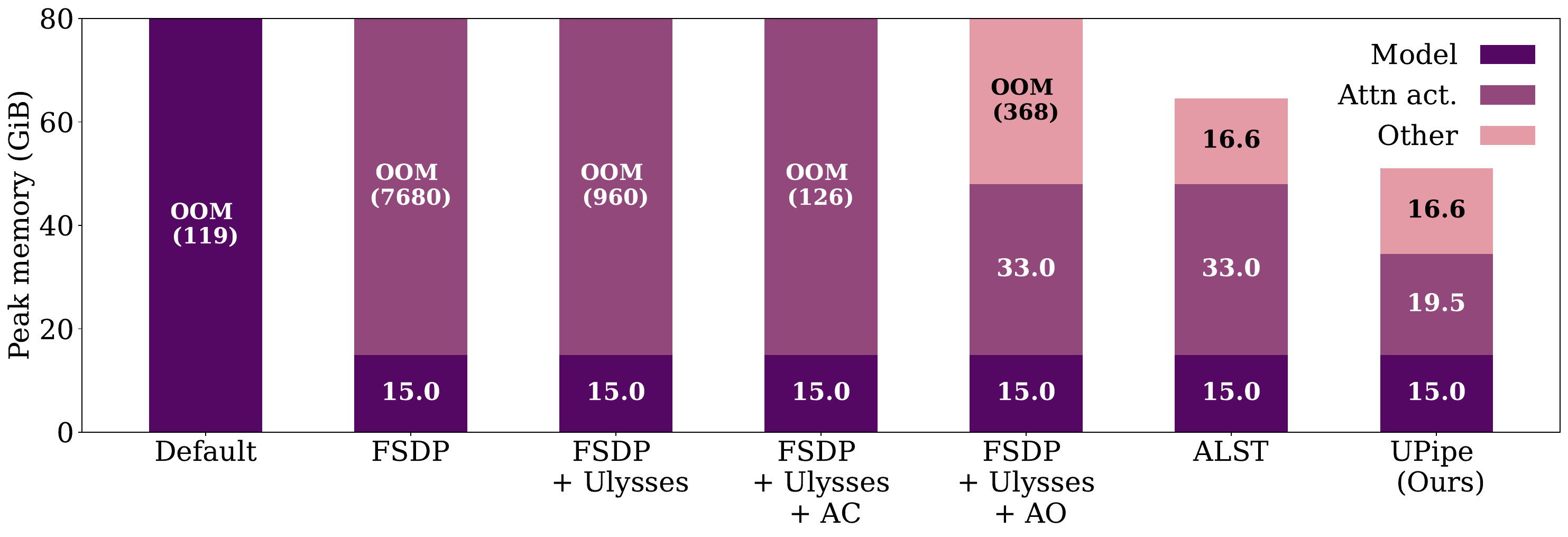}
    \vspace{-4ex}
    \caption{Memory usage breakdown when training Llama 3-8B with a sequence length of 3M tokens across 8 H100 GPUs. FSDP denotes Fully Sharded Data Parallel,
        AC denotes Activation Checkpointing, AO denotes AC with offloading, OOM denotes Out of Memory.}
        \vspace{-2.0ex}
        \label{fig:stacked_mem_chart}
\end{figure*}

In this paper, we propose \textbf{\ours{}}, a context parallelism method that focuses on improving the memory efficiency of long-context training while maintaining performance on par with current approaches. 
Our method is designed on the principle that for long-context training, processing a subset of heads at once is enough to saturate the GPU.
Therefore, serializing the execution in the attention layer by grouping heads into smaller chunks allows for much better memory reuse.
\ours{} is agnostic to the underlying attention algorithm: similar to DeepSpeed Ulysses, it uses the same kernels to compute attention as non-distributed training.

On Llama 3-8B, our method can fit context lengths of \textbf{up to 5 million tokens on a single H100 node}, and up to 8M on two H100 nodes with unified sequence parallelism~\cite{usp}, outperforming prior works by \textbf{25\%} and \textbf{33\%} respectively in terms of maximum context length supported.
At the same time, \ours{}'s training throughput is comparable to other context parallelism techniques.

Our contributions are as follows:
\vspace{-1em}
\begin{enumerate}
    \item We propose \ours{}, a new context parallelism method that executes the attention layer in multiple stages, processing attention heads in chunks. 
    This method is easy to implement and can work as a plug-and-play replacement for existing techniques.
    \item We analyze the memory usage of long-context Transformer training, identifying the major bottleneck left unaddressed by prior works and showing how \ours{} mitigates this bottleneck.
    \item We present a schedule compatible with Grouped-Query Attention (GQA), which processes heads out of order to avoid redundant communication while retaining the memory benefits of standard GQA architecture.
    \item We compare \ours{} with prior approaches to context-parallel training, measuring the speed and memory usage for 8B and 32B dense Transformer models across sequence lengths ranging from 128K to 5M tokens\footnote{The code for our experiments is available at \href{https://github.com/ghadiaravi13/Untied-Ulysses}{\texttt{github.com/ghadiaravi13/Untied-Ulysses}}.}. 
    \ours{} can support longer sequence lengths (up to 5 million tokens on a single 8$\times$H100 node) with negligible performance differences, compared to other methods.
\end{enumerate}

\begin{table*}[t]
    \centering

    \caption{Theoretical peak memory usage breakdown across different stages of the forward pass for a Transformer model. All floating point tensors use \texttt{BFloat16} precision by default, except cross-entropy loss, which uses \texttt{fp32} precision.}
\label{tab:tensor_mem}
    \renewcommand{\arraystretch}{1.0}
    \resizebox{1.0\textwidth}{!}{%
    \begin{tabular}{@{}cc c c c c c c@{}}
    \toprule
    & \multirow{2}{*}[-2pt]{\textbf{Stage}} &
    \multirow{2}{*}[-2pt]{\textbf{Inputs}} &
    \multicolumn{3}{c}{\textbf{Intermediate tensors}} &
    \multirow{2}{*}[-2pt]{\textbf{Outputs}} &
    \multirow{2}{*}[-2pt]{\textbf{Total}} \\ \cmidrule(lr){4-6}
     & & & \textbf{Type} & \textbf{Memory} & \textbf{Ratio} & & \\ \midrule
    
    \ballnumber{1} & Embedding &
    $4\cdot S$ (\texttt{int32}) &
    -- & -- & -- &
    $2\cdot S \cdot d_{model}$ &
    $2\cdot S \cdot d_{model}$ \\ \midrule
    
    \ballnumber{2} & Attention &
    $2\cdot S \cdot d_{model}$ &
    \begin{tabular}[c]{@{}c@{}}QKV \\ all-to-all \\ \end{tabular} & 
    \begin{tabular}[c]{@{}c@{}}$6\cdot S \cdot H \cdot d_{head}$ \\ $6\cdot S \cdot H \cdot d_{head}$\\ \end{tabular} & 
    \begin{tabular}[c]{@{}c@{}}$H = d_{model}/d_{head}$ \end{tabular} &
    $2\cdot S \cdot d_{model}$ &
    \begin{tabular}[c]{@{}c@{}} $16\cdot S \cdot d_{model}$ \\ \end{tabular} \\ \midrule
    
    \ballnumber{3} & Feed-forward &
    $2 \cdot S \cdot d_{model}$ &
    Intermediate &
    $8\cdot S \cdot d_{ff}$ &
    $d_{ff} \approx 2.67\cdot d_{model}$ &
    $2\cdot S \cdot d_{model}$ &
    $25\cdot S \cdot d_{model}$ \\ \midrule
    
    \ballnumber{4} & Cross-entropy &
    $2 \cdot S \cdot d_{model}$ &
    Logits + LogSoftmax &
    $8\cdot S \cdot V$ &
    $V \approx 30\cdot d_{model}$ &
    Loss &
    $240\cdot S \cdot d_{model}$ \\ \bottomrule
    \end{tabular}%
    }

\end{table*}

\section{Background}

Long-context Transformer training is increasingly important for use cases such as voice~\cite{speech} and video generation~\cite{magi1}, but standard self-attention scales poorly with sequence length. 
For instance,~\citet{wan} report that the activation memory for training a 14B Diffusion Transformer at a sequence length of 1M tokens reaches 8 TB, far beyond the capacity of a single accelerator and necessitating distributed training.

\vspace{-1ex}

\subsection{Multi-head Attention}

\vspace{-0.5ex}

Multi-head self-attention~\cite{transformer} is a neural network layer for processing sequential data. 
For head $h$, each input vector $X_i$ is projected into query $Q^h_i$, key $K^h_i$, and value $V^h_i$ vectors, and the output is computed as $\textrm{softmax}\left(\frac{Q^h_i \cdot K^h_{\leqslant i}}{\sqrt{d_h}}\right) \cdot V^h_{\leqslant i}$, where $d_h$ is the query vector size. 
Since each query attends to \textit{all} past keys and values, computing attention requires access to the entire sequence, which becomes the key obstacle for long-context training.
Grouped-Query Attention (GQA,~\citealp{gqa}) shares the same key/value heads among groups of $G$ query heads, lowering the key/value memory use by a factor of $G$.

\vspace{-1ex}

\subsection{Context Parallelism}

\vspace{-0.5ex}

To mitigate the large activation memory footprint, \citet{ring} introduced context parallelism in the form of Ring Attention. 
It shards the context across $C$ devices and exchanges $K, V$ tensors in a ring via peer-to-peer communication. 
This incurs $\mathcal{O}(C)$ communication calls per attention operation. 
Later, \citet{ring_new} proposed a variant of Ring Attention that is compatible with online softmax computation.
DeepSpeed-Ulysses~\cite{ulysses}, built upon Megatron-SP~\cite{megatron-sp}, instead rearranges tensors via a single all-to-all collective, reducing communication latency while unlocking optimized self-attention kernels. 
We discuss this technique in detail in Section~\ref{sec:ds_ulysses}.

Further works introduce additional optimizations for memory and throughput. 
For instance, USP (Unified Sequence Parallelism,~\citealp{usp}) uses DeepSpeed-Ulysses within the node and Ring Attention across nodes, running all-to-all communication over the faster NVLink fabric and ring communication over the slower one. 

To address the memory bottleneck of very long contexts, other papers have explored tiling or chunking techniques. 
In particular, Arctic Long Sequence Training (ALST,~\citealp{alst}) uses tiling for feed-forward layer and cross-entropy loss calculations, lowering the memory pressure of intermediate tensors.
However, it does not address the memory overhead in the attention phase. 
Fully Pipelined Distributed Transformer (FPDT,~\citealp{fpdt}) addresses the attention memory overhead by chunking computations along the sequence length dimension and offloading to CPU.
However, it suffers from reduced performance due to CPU overhead and additional memory transfers. 
\ours{} addresses the memory overhead while maintaining performance and extends to hybrid schemes such as USP. 
Also, our technique performs chunking along the head dimension, which is orthogonal and complementary to FPDT.

\vspace{-1.0ex}

\subsection{Activation memory for a Transformer model}

\label{act-mem}

Next, we quantify and analyze the memory usage in each phase of a long-context Transformer training step.
We identify the factors that impose the biggest memory bottlenecks and cover the techniques to address them.

Consider a decoder-only Transformer with $L$ layers, with $H$ heads per layer, and a GQA group size of $G$. 
Let the model's hidden size be $d_{model}$, per-head dimension $d_{head}$, the intermediate dimension of the feed-forward network $d_{ff}$, and the vocabulary size of the model $V$. 
Finally, let a sequence of length $S$ be the input to the model. 
For now, we keep the batch size equal to 1, since our focus is to establish how memory scales with respect to the sequence length $S$. 

We assume a mixed precision setup with parameters, activations, and gradients in \texttt{bfloat16} precision, thus requiring 2 bytes per parameter for each tensor. 
However, some intermediate tensors still require 4 bytes, as discussed later.

For a decoder-only Transformer model, the sequence of operations during the forward pass can be broken down into 4 phases, as shown in Table \ref{tab:tensor_mem}.


\ballnumber{1} \textbf{Embedding:} The input sequence of tokens is converted into embedding vectors of size $ S \cdot d_{model} $, thus requiring $2\cdot S \cdot d_{model} $ bytes of memory.

\vspace{0.25ex}

\ballnumber{2} \textbf{Attention:} The input gets transformed into query $Q$, key $K$ and value $V$ vectors, requiring $6\cdot S\cdot H\cdot d_{head} $ bytes.
As we will discuss in Section \ref{sec:ds_ulysses}, all-to-all requires the same amount of additional memory.
Assuming we use Flash Attention \cite{flashattn}, the attention computation would not require any additional GPU HBM memory, except for storing the final output (and the final log-sum-exponent) vectors of size $2\cdot S\cdot d_{model} \space$ (and $2\cdot S\cdot H $ respectively). 
For most Transformer models, $H = d_{model}/d_{head}$; therefore, this phase has a total memory usage of $\label{stage2-mem}2+(6+6)+2 = 16\cdot S\cdot d_{model} $ bytes.

\vspace{0.25ex}

\ballnumber{3} \textbf{Feed-forward:} Next, the attention output is fed to the feed-forward network (FFN) with two layers. 
Assuming SwiGLU activations \cite{swiglu}, this module projects the input with dimension $d_{model}$ into four intermediate tensors of dimension $d_{ff}$ and then projects the intermediate tensors into the output of dimension $d_{model}$. 
Typically, $d_{ff} \approx 2.67\cdot d_{model}$, which results in the total memory usage of $25\cdot S\cdot d_{model}$ bytes.

\vspace{0.25ex}

\ballnumber{4} \textbf{Cross-entropy loss:} At the end of the final layer, the output is converted into logits of shape $S\cdot V$, which are then used by the cross-entropy loss function. 
This phase has the most critical memory constraints, because the vocabulary size $V$ is typically very large ($\approx 30\cdot d_{model}$).
Moreover, the cross-entropy calculation requires logits and intermediate log-softmax values to be casted to \texttt{fp32}, making the overall memory consumption $240\cdot S\cdot d_{model}$ bytes. 


\begin{table*}[t]
    \centering
    \caption{Peak activation memory within the forward attention block under GQA, expressed in units of one local hidden-state shard ($\frac{S}{C}\cdot d_{model}$ \texttt{bf16} elements, i.e., $2\cdot\frac{S}{C}\cdot d_{model}$ bytes). $\pi$ represents the number of chunks in FPDT, $\nu$ represents the number of chunks in UPipe.
    UPipe consumes $\nu$ times less intermediate (QKV + all-to-all) activation memory than Ulysses with activation offloading. FPDT has lower memory usage due to an arbitrary chunk size, but suffers from performance degradation.}
    \label{tab:gqa_fwd_mem}
    \vspace{-1.0ex}
    \begin{tabular}{@{}lcccc@{}}
        \toprule
        \textbf{Method} & \textbf{Before attn block} & \textbf{During inp\_all\_to\_all} & \textbf{During attn kernel} & \textbf{During out\_all\_to\_all} \\
        \midrule
        Ulysses & $L$ & $L + (\gamma+1)$ & $L + (\gamma+1)$ & $L + 2$ \\
        \midrule
        Ulysses + offloading & $1$ & $\gamma + 2$ & $\gamma + 2$ & $3$ \\
        \midrule
        FPDT & $\frac{1}{\pi}$ & $\frac{\gamma+2}{\pi}$ & $\frac{2\gamma+1}{\pi}$ & $\frac{2}{\pi}$ \\
        \midrule
        Untied Ulysses & $1$ & $2 + \frac{\gamma}{\nu}$ & $2 + \frac{\gamma}{\nu}$ & $2 + \frac{1}{\nu}$ \\
        \bottomrule
    \end{tabular}%
    \vspace{-2.5ex}
\end{table*}



\subsection{Mitigating memory overheads}

From the above analysis, we can identify the most critical factors that become a memory bottleneck when scaling the context length. 
Due to its memory usage, the cross-entropy loss is the first obstacle.
We can overcome it by computing the loss in a tiled manner, materializing the intermediate tensors only one tile at a time.
To reduce the memory usage of the FFN layer, we adopt a similar approach to ALST by using a tiled FFN function.
Similarly to other prior works and libraries like Unsloth~\cite{unsloth}, we manage the activation memory across layers via full activation checkpointing with CPU offloading.
Finally, for the attention layer, we propose \textbf{\ours{}}, described in Section~\ref{untying ulysses}.

Table \ref{tab:gqa_fwd_mem} provides the memory consumption of different stages of attention during the forward pass. 
In grouped-query attention, the size of key ($K$) and value ($V$) tensors is reduced by a factor of $G$. 
We define $\gamma = 1 + \frac{2}{G}$ as the combined size of $Q, K, V$ (relative to $S/C\cdot d_{model}$). 
Table \ref{tab:gqa_bwd_mem} of Appendix \ref{app:a} presents a similar breakdown for the backward pass of an attention layer (with grouped-query attention).

\vspace{-1.0ex}

\begin{figure*}[h] 
    \centering
    \includegraphics[width=1.0\textwidth]{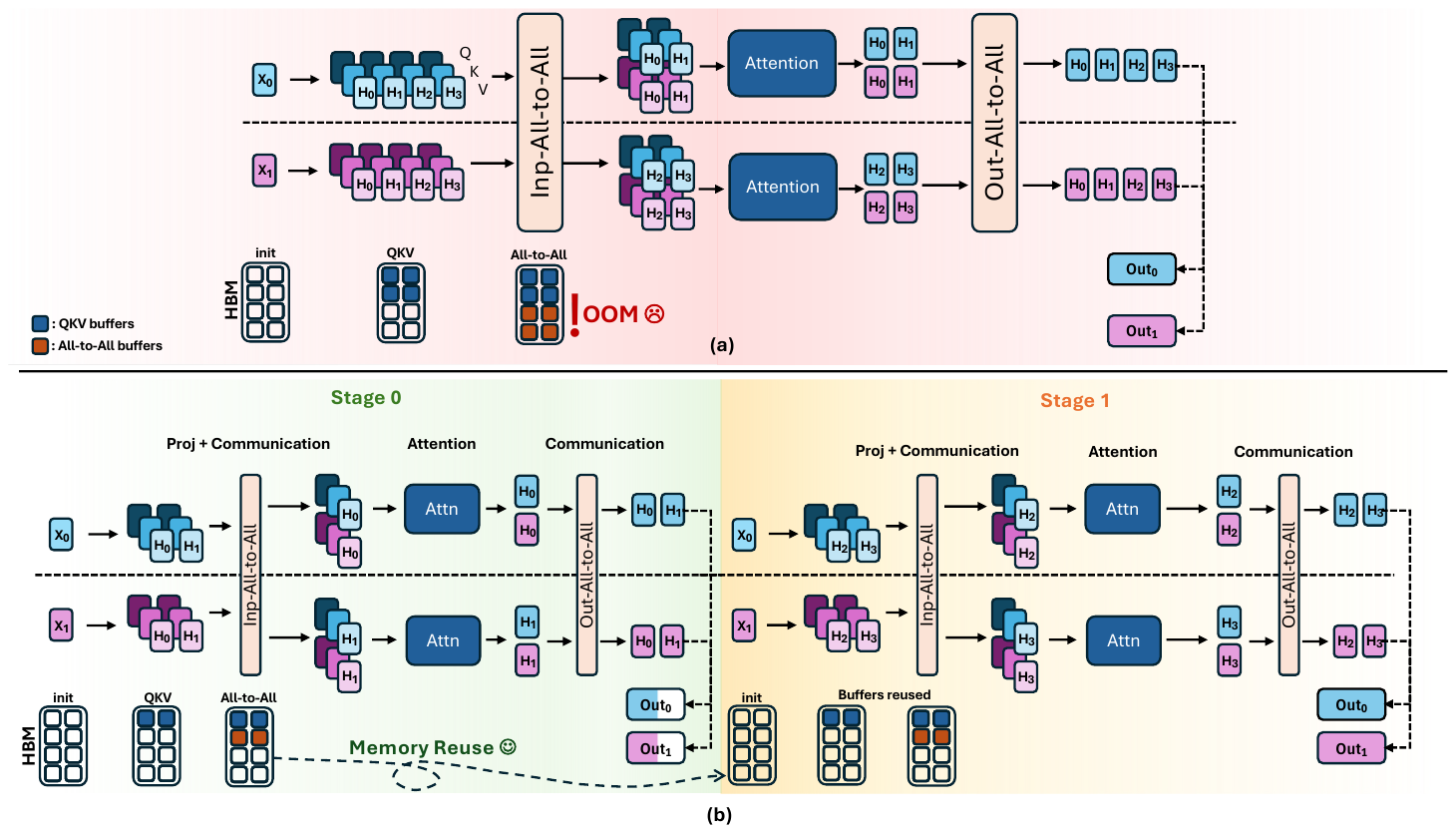}
    \caption{Illustration of (a) DeepSpeed-Ulysses and (b) \ours{} designs. \ours{} processes attention in a headwise \textit{untied} manner, so that in each stage, attention is performed only on a subset of heads. 
    This allows memory reuse across different stages, significantly reducing the peak memory usage due to attention activations. 
    HBM (High-Bandwidth Memory) usage illustrates memory utilization of the intermediate buffers and omits other components for brevity.}
    \label{fig:main_fig}
\end{figure*}


\section{Untied Ulysses}

\vspace{-0.5ex}

As discussed above, the majority of activation tensors during training that contribute to peak memory usage can be tiled, recomputed or offloaded. 
In this section, we outline the design of \ours{}, highlighting its key ideas that enable memory savings in the attention stage. 
Since our method builds on top of DeepSpeed-Ulysses~\cite{ulysses}, we will first elaborate on the mechanics of this algorithm, highlighting the opportunity for memory optimization.

\vspace{-1.0ex}

\subsection{DeepSpeed-Ulysses}

\vspace{-0.5ex}

\label{sec:ds_ulysses}

DeepSpeed-Ulysses (DS-Ulysses) is a context parallelism technique that allows training of Transformer models in a distributed setup by sharding input sequences across devices.
For a sequence of length $S$ and $C$ devices, every device will have a sequence of length $S/C$. 
Token-wise operations (feed-forward layer, RMSNorm, cross-entropy) can be applied to every sequence shard without communication. 
However, the attention layer requires access to the entire sequence, so it cannot be executed independently on each shard. 
Ulysses solves this problem by rearranging the shards: the memory occupied by the shards is the same as before, but now every device has access to the full sequence.

Let us walk through how DeepSpeed-Ulysses works, using an example from Figure \ref{fig:main_fig}a with $C = 2$ devices. 
Initially, the sequence $X$ of size $[S,d_{model}]$ is sharded on two devices: $X_0$ and $X_1$ with size $[\frac{S}{2},d_{model}]$ each. 
These shards then undergo the $QKV$ projection to produce the query, key, and value tensors. 
In this example, let us consider the number of heads $H$ = 4, which means the query, key, and value tensors have 4 heads: $H_0, H_1, H_2, H_3$ with $d_{head}=d_{model}/4$.
Hence, the size of these tensors is $[\frac{S}{2},4,d_{head}]$. 

Next, DS-Ulysses performs an all-to-all operation \textbf{inp\_all\_to\_all} on the $QKV$ tensors to \textit{reshard} them: switching the sharding from the sequence length dimension to the head dimension. 
Thus, $QKV$ tensors are resharded from $[\frac{S}{2},4,d_{head}]$ to $[S, 2, d_{head}]$. 
As a result, device 0 now has access to the full sequence for heads $H_0$, $H_1$, while device 1 has access to the full sequence for heads $H_2$ and $H_3$. 
After attention, device 0 computes outputs for heads $H_0$ and $H_1$ (and device 1 computes outputs for $H_2$ and $H_3$), which are then sharded back to the original shape. 
Therefore, another all-to-all operation \textbf{out\_all\_to\_all} is performed, so that the final outputs $O_0$ and $O_1$ are of shape $[\frac{S}{2},d_{model}]$ each.

Because of all-to-all communication, DS-Ulysses can fully utilize the network topology between all context-parallel ranks and does not incur a latency overhead of $\mathcal{O}(C)$ peer-to-peer transfers, which makes it more performant than Ring Attention~\cite{loongtrain}. 
However, it faces a memory bottleneck for long-context training, because the overhead of communication buffers can become significant.


\subsection{Memory Usage}
\label{mem_usage_section}
The memory usage peak in DeepSpeed-Ulysses occurs during the \textbf{inp\_all\_to\_all} operation because of the memory buffers for query, key, and value tensors and similarly sized all-to-all buffers.
From Table \ref{tab:tensor_mem}, we can see that the memory usage due to these intermediate tensors is proportional to $H$. 
Our key insight is that \emph{for long sequences and large enough models, performing attention over a subset of heads is enough to reach the compute-bound regime}. 
Hence, we process only $U$ heads at a time, where $U<H$.
Because the memory usage of intermediate tensors is proportional to the number of heads, the memory overhead reduces from $\mathcal{O}(H)$ to $\mathcal{O}(\text{$U$}).$ 
As we show later, changing the hyperparameter $U$ allows for a natural runtime-memory tradeoff.

\begin{figure*}[ht] 
    \centering
    \includegraphics[width=1.0\textwidth]{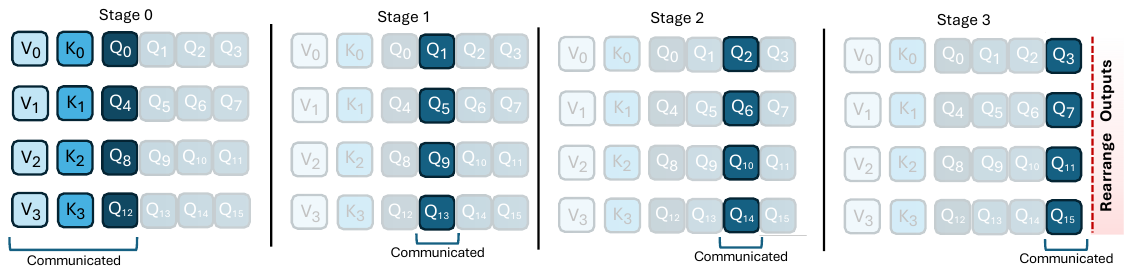}
    \caption{Illustration of \ours{}'s GQA scheduling algorithm. We communicate as many unique key/value heads as possible, along with the corresponding queries in stage 0. In the subsequent stages, we only communicate the next queries of the corresponding groups, reusing the key/value tensors from stage 0 until stage $G$, where $G$ is the group size.}
    \label{fig:scheduling}
    \vspace{-7pt}
\end{figure*}

\subsection{Untied Ulysses: \ours{}}

\vspace{0.75ex}

\label{untying ulysses}

The biggest memory overhead in DS-Ulysses comes from storing the QKV tensors and corresponding all-to-all buffers \textit{for all heads}. 
\ours{} addresses this by untying the entire attention execution in an end-to-end manner. We propose a headwise chunking scheme for performing attention, which processes only a subset of heads at a time, reducing the peak activation memory. To realize the memory benefits in practice, we also propose a GQA-scheduling technique that processes attention heads out-of-order and prevents redundant communication of key and value heads.

\vspace{0.75ex}

Formally, given a model with attention heads $H$ and number of context parallel devices $C$, \ours{} chunks the attention execution into $H/U$ stages, processing $U$ heads per stage.
During the forward pass, the execution begins by projecting the input $X$ into $Q_{U}^{0}$, $K_{U}^{0}$, and $V_{U}^{0}$, i.e., the first $U$ heads. 
This is followed by \textbf{inp\_all\_to\_all}, resulting in $U/C$ heads per device. Note that $U$ must be divisible by $C$ to ensure that each device processes an integer number of heads. 

\vspace{0.75ex}

In the next stage, we process the next $U$ heads while reusing the memory buffers from the previous stage (i.e., use $Q_{U}^{0}$ buffers to store  $Q_{U}^{1}$ and similarly for other tensors). Thus, the memory usage remains $\mathcal{O(\text{$U$)}}$ throughout the execution.

Let us now walk through the example in Figure \ref{fig:main_fig}b with $H=4$, $U=2$, and $C=2$. \ours{} performs attention over $H/U=2$ stages: processing 2 heads per stage. 

Consider the execution on \textbf{Device 0}: in stage 0, the input $X_0$ is projected into the first two heads $H_0$, $H_1$ of $QKV$. 
Next, \textbf{inp\_all\_to\_all} is performed on $H_0$, $H_1$ so that device 0 has $H_0$ for the entire sequence.
Notice that during all-to-all, we only need buffers for 2 heads (as opposed to 4 heads in DS-Ulysses). 
We then perform attention on $H_0$, generate output for head 0, and perform \textbf{out\_all\_to\_all}. 
In the second stage, $X_0$ is projected into the next two heads $H_2$, $H_3$. 
At this point, heads $H_0$, $H_1$ are already processed, so we reuse their HBM buffers to store $H_2$, $H_3$. 
Next, we perform all-to-all and similarly reuse the all-to-all buffers from stage 0. 
For the final output, we initialize the buffers at the beginning and fill them during execution. 
This avoids the concatenation of individual output chunks at the end of the attention stage, which otherwise degrades performance.

\vspace{-1.5ex}

\subsection{Memory Savings}

For $H$ heads and $C$ devices used for context parallelism, DS-Ulysses has a total memory usage of: $6\cdot \frac{S}{C}\cdot H\cdot  d_{head}$ bytes for the $QKV$ tensors and the same number of bytes for the all-to-all communication buffers, resulting in a total of $12\cdot \frac{S}{C}\cdot H\cdot d_{head}$ bytes of intermediate tensors.

In case of \ours{}, we can replace $H$ with $U$, because a single stage processes $U$ heads. 
Thus, the memory usage is $12\cdot \frac{S}{C}\cdot U\cdot d_{head}$ bytes. 
To maximize memory savings, we want $U$ to be as small as possible, and the smallest valid value is $U=C$.
In this setting, the maximum memory usage of \ours{} during the attention stage becomes $12\cdot S\cdot d_{head}$.
Hence, when $U=C$, the peak activation memory usage of our method is independent of the number of heads.

Taking Qwen3-32B as an example with $H=64$ and using a single $8\times$H100 node (i.e., $C=8$), DS-Ulysses requires $96\cdot S\cdot d_{head}$ bytes of memory. 
By contrast, \ours{} requires $12\cdot S\cdot d_{head}$ bytes of memory, lowering the memory usage due to intermediate activations for self-attention by \textbf{87.5$\%$}.

\vspace{-1.5ex}

\section{Implementation}

We use TorchTitan~\cite{torchtitan} as the training framework to integrate the implementation of \ours{} due to its flexibility and performance.
We use the same framework for Unified Sequence Parallelism (USP) baselines to ensure a fair comparison across different CP techniques. 

We integrate the optimized layers inside TorchTitan via a simple drop-in replacement of the existing ones. 
As the baseline for Ulysses and Ring Attention, we use the hybrid attention implementation from USP, which includes a load-balanced zig-zag Ring Attention implementation. 
UPipe uses the same code structure as USP, so it can be easily extended for a hybrid setup (Ulysses+Ring). 
For all-to-all communication, we use the non-QKVPacked variant from USP, which transfers queries, keys, and values sequentially to avoid memory overhead of simultaneous transfers.

For operations independent along the sequence length, such as FFN and Root Mean Squared Normalization (RMSNorm), we use TiledCompute --- a tiling mechanism introduced by ALST~\cite{alst}. 
We use tiling for RMSNorm, since we found it to be more memory-efficient than using \texttt{torch.compile}. 
Similarly to the authors of ALST, we use a square tile of size $[d_{model}\times d_{model}]$. 
We use \texttt{FusedLinearCrossEntropyLoss} from Liger-Kernel~\cite{liger} for memory-efficient loss computation. 
This kernel fuses the final linear projection with the cross-entropy loss, computing logits and loss in a tiled manner to avoid materializing the full \texttt{fp32} logits tensor in memory.
We also find that Rotary Positional Encoding \cite{rope} also incurs a memory overhead due to \texttt{fp32} casting.
Hence, we use the fused RoPE implementation from the Flash Attention API, which performs in-place operations to avoid large memory spikes.


\subsection{Grouped-Query Attention Scheduling}
\label{gqa-scheduling}

\vspace{0.5ex}

Because of its memory benefits, GQA is an important architectural component of modern Transformer models~\cite{llama3,mistral,qwen3}. 
To retain this advantage, we make our design GQA-compatible. 
This requires rearranging the query tensors to reuse the KV tensors communicated in the previous stage.

Assuming $U=C$, recall that UPipe processes $C$ heads per stage. 
For a GQA model with group size $G$, only $C/G$ of the first $C$ heads are unique. 
For example, consider $C$ = 4 and $G$ = 4. 
With naive processing, stage 0 would process query heads $Q_0$, $Q_1$, $Q_2$, $Q_3$ with the corresponding key heads $K_0$, $K_0$, $K_0$, $K_0$. 
Since each device transfers $C-1$ query, key, and value heads per stage, the total communication volume across $H/C$ stages is $\mathcal{O(\text{3} \cdot \frac{\text{$H$}}{\text{$C$}}\cdot {\text{$(C-1)$}})}=\mathcal{O}(3H)$.

\vspace{0.5ex}

As shown in Figure \ref{fig:scheduling}, we could communicate $K_0$, $K_1$, $K_2$, $K_3$ (i.e., all the unique key tensors) in stage 0.
We would then need to send the corresponding query tensors (i.e., $Q_0$, $Q_4$, $Q_8$, $Q_{12}$). 
However, in the next stage, we do not need to communicate any key (or value) tensors.
We can now choose the next query tensors from the groups (i.e., $Q_1$, $Q_5$, $Q_9$, $Q_{13}$), because the corresponding key (and value) tensors were already communicated in stage 0. 
In this case, for every $G$ stages, we communicate $C-1$ query, key, and value heads in the first stage, and only $C-1$ query heads in the following stages. 
Thus, the total communication volume is $\mathcal{O\left(\text{$(3+G-1)$} \cdot \frac{\text{$H$}}{\text{$C\cdot G$}}\cdot \text{$(C-1)$}\right)}=\mathcal{O}\left(\frac{2H}{G}+H\right)$, which is always less than the naive processing, because $G > 1$.

\section{Experiments}

\begin{table*}[h]
\centering
\caption{\textbf{Throughput comparison} (tokens/second/GPU) for \textbf{Llama 3-8B} (8$\times$H100s) and \textbf{Qwen3-32B} (16$\times$H100s) across varying sequence lengths.  
On a single 8$\times$H100 node, our method scales Llama 3-8B to a 5M-token sequence length, reaching a 25$\%$ improvement over the previous state-of-the-art. \textbf{OOM}: Out of Memory. \textbf{Note}: FPDT execution fails at lengths $>$ 4M.}
\label{tab:performance}
\begingroup
\tiny 
\resizebox{\textwidth}{!}{
\begin{tabular}{@{}l@{}l@{}cccccccccc}
\toprule
\vspace{1pt}
& \phantom{xxxx}Method & 128K& 256K& 512K& 1M& 2M& 3M& 4M& 5M\\
\midrule 

\multirow{5}{*}[0em]{\rotatebox[origin=c]{90}{Llama 3-8B}}

& \phantom{x}Native PyTorch\phantom{x}& 1373.87& 845.99& 474.30& 249.85& OOM& -- & -- & --\\
& \phantom{x}Ring& 2064.90& 1387.67& 841.05& 458.51& 237.99& 159.96& OOM & --\\
& \phantom{x}Ulysses&\textbf{2320.47}& \textbf{1503.80}& \textbf{878.63}& \textbf{475.33}& 246.05& 162.41 & OOM & --\\
& \phantom{x}FPDT& 1171.68& 884.75& 621.20& 382.42& 219.53& 153.48& 119.76 & --\\
& \phantom{x}\ours{}& 2281.05& 1487.29& 867.17& 472.53& \textbf{246.07}& \textbf{166.32}& \textbf{125.56} & \textbf{98.25} \\
\midrule

\multirow{5}{*}[0em]{\rotatebox[origin=c]{90}{Qwen3-32B}}

& \phantom{x}Native PyTorch\phantom{x}& 127.03& 112.20& 91.39& OOM& -- & -- & -- & --\\
& \phantom{x}Ring& 418.39& 308.88& 194.44& 110.27 & 58.45 & OOM & -- & --\\
& \phantom{x}Ulysses&\textbf{545.29}& \textbf{370.70}& \textbf{217.04}& \textbf{117.02}& \textbf{59.98}& OOM & -- & --\\
& \phantom{x}FPDT& 286.40 & 217.85& 151.91& 95.88& 55.41& 38.86& 27.66 & --\\
& \phantom{x}\ours{}& 483.29& 339.56& 204.46 & 113.26 & \textbf{59.56} & \textbf{40.42}& \textbf{29.97} & OOM\\

\bottomrule
\end{tabular}
}
\endgroup
\end{table*}

We now present the experimental results using \ours{}, comparing it with multiple state-of-the-art baselines, as well as standard context parallelism approaches. 
We run the experiments on two models across multiple context lengths, comparing the maximum context length supported by different techniques and their throughput in each setup. 
Unless stated otherwise, for all experiments with \ours{}, we match the chunk size with the number of context-parallel devices ($U$ = $C$) to reach the highest memory efficiency.

\subsection{Setup}
\label{subsec:setup}

We run our experiments on $8\times$ NVIDIA H100 nodes with 80GiB of on-chip DRAM per GPU, connected via 4th generation NVLink with 900GBps bidirectional bandwidth for intra-node communication and 400 Gbps Infiniband networking for inter-node communication. Each node has a Intel Xeon Platinum 8480+ CPU with 1.9TiB RAM.

We use Llama 3-8B~\cite{llama3} and Qwen3-32B~\cite{qwen3} models in our experiments. Llama 3-8B has $H=$ 32 query heads (and 8 key-value heads), allowing \ours{} to process attention in $H/C = $ 4 stages. 
Qwen3-32B has 64 query heads (and 8 key-value heads), so \ours{} processes it in 8 stages. Note that we use $C = $ 8 here, because we always restrict the Ulysses context parallelism degree to 8 and use the remaining mesh for Ring Attention in a hybrid style, as discussed in Section~\ref{sec:baseline}.

Our experiments use TorchTitan and Flash Attention-3 (FA3, ~\citealp{fa3}). 
We use context parallelism over 8$\times$H100 nodes for our training. 
We use PyTorch's FSDP2 (natively supported by TorchTitan) to distribute parameters, gradients, and optimizer states across all GPUs.

To manage activation memory efficiently and maintain consistency with FPDT, we use full activation checkpointing with CPU offloading. 
For all sequence lengths except 5M, we allow the offloaded activations to reside in the non-swappable CPU RAM by setting \texttt{pin\_memory} to True. 
For the 5M context, we set this to False due to the CPU RAM constraints. 
For better memory management, we set \texttt{PYTORCH\_CUDA\_ALLOC\_CONF} \texttt{=expandable\_segments:True}, similarly to ALST.


\subsection{Baselines}
\label{sec:baseline}

We compare the performance and memory efficiency of UPipe against multiple baselines. 
Fully Pipelined Distributed Transformer (FPDT, \citealp{fpdt}) is a long-context training method that processes attention by chunking along the sequence length dimension and using online softmax to perform full attention. 
It mitigates the memory requirements by asynchronously offloading chunks to the CPU, keeping only the necessary chunks on the GPU, and using a double buffer mechanism to overlap communication with computation. 
It can support a maximum sequence of 4M tokens when training Llama 3-8B on a single 8$\times$H100. 
The original implementation of FPDT does not natively support Flash Attention 3 or SwiGLU activations, which is why we patched it for a fair comparison with \ours{}.

For standard Ulysses and Ring Attention baselines, we use the USP implementation, as it provides an easy-to-use interface for modifications. 
Additionally, we also compare with the native Ring Attention implementation from PyTorch. 
Our experiments omit ALST, because our modified version of USP-Ulysses that uses DS-Ulysses, offloaded activation checkpointing, and tiling for MLP / CELoss integrates all improvements from the ALST design. 
However, unlike ALST, we do not offload optimizer states to the CPU to avoid throughput degradation.

For multi-node experiments, we use a similar setup as USP-Hybrid with \textit{8-ulysses-2-ring}, which denotes using Ulysses over 8 GPUs within the node and Ring Attention across 2 nodes. 
This is a common setup (e.g.,~\citealp{wan}) to allow faster all-to-all communication for Ulysses within the node, and slower ring communication across nodes. 
Note that FPDT does not support hybrid context parallelism, so we use the standard \textit{16-ulysses-1-ring} setup. 

We also compare against pure Ring Attention baselines: USP-Ring and a native PyTorch implementation. Both of them use zig-zag load balancing, ensuring fair work distribution across all ranks participating in context parallelism.

\subsection{Performance}

\vspace{0.5ex}

\subsubsection{Single-Node Training}

\textbf{Llama 3-8B:} Table \ref{tab:performance} (top) reports single-node throughput for Llama 3-8B training across increasing sequence lengths.
Ulysses has high throughput due to a single all-to-all call per attention, but its full-head $QKV$ and communication buffers make activation memory the primary limiter at multi-million token contexts.
FPDT can push context length further, but it pays a substantial throughput penalty from frequent CPU--GPU transfers due to fine-grained CPU offloading.

\vspace{0.5ex}

\ours{} bridges this gap: at shorter sequences, it is slightly slower than Ulysses due to additional stage launches, but this overhead is amortized with longer contexts (as shown in Table \ref{tab:runtime}).
At longer sequences (\,$\geq$\,2M), \ours{} matches Ulysses throughput while using much lower GPU memory, enabling training with 5M tokens on a single 8$\times$H100 node.
In effect, \ours{} preserves the throughput of Ulysses while having the memory efficiency closer to offloading-based methods, improving the maximum single-node context length over FPDT (4M tokens) by \textbf{25\%}.

\begin{figure}[t]
    \centering
    \includegraphics[width=1.0\columnwidth]
    {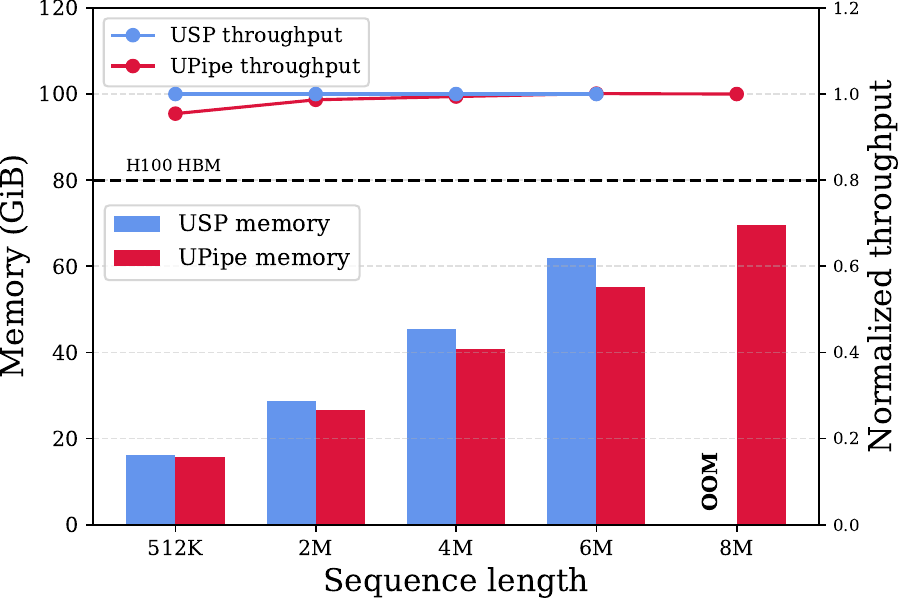}
    \vspace{0.5ex}
    \caption{\textbf{Llama 3-8B:} Comparison of peak GPU memory usage and throughput (normalized w.r.t. USP-Hybrid) between \ours{} and USP-Hybrid at different sequence lengths on 16$\times$H100s.}
    \label{fig:multinode-mem}
    \vspace{-3.0ex}
\end{figure}

\subsubsection{Multi-Node Training}


\textbf{Llama 3-8B:} Figure \ref{fig:multinode-mem} shows the comparison of \ours{} and USP-Hybrid on a 16$\times$H100 setup. 
\ours{} is more memory efficient than USP-Hybrid at all sequence lengths from 512K to 6M tokens, and supports context lengths up to 8M tokens, improving upon USP-Hybrid by \textbf{33\%}. 
The throughput of UPipe is comparable to USP-Hybrid at all sequence lengths, highlighting \ours{}'s runtime efficiency.


\textbf{Qwen3-32B:} Table \ref{tab:performance} (bottom) summarizes the throughput comparison of different methods when training Qwen3-32B on a 16$\times$H100 setup. 
\ours{} outperforms all other methods for sequences $\geq$ 2M. 
Notably, \ours{} always outperforms FPDT across all sequence lengths in terms of throughput. 
In terms of maximum sequence length, our method can support 4M token sequences, 2$\times$ more than Ulysses (2M tokens), while delivering \textbf{8.3\%} better performance than FPDT.

Additionally, we also report the maximum allocated memory in Table~\ref{tab:appendix-performance} of Appendix \ref{app:b}. \ours{} provides the best memory efficiency compared to other methods except FPDT. However, FPDT suffers from performance degradation due to frequent CPU transfers. Nevertheless, our method should be composable with FPDT due to orthogonal chunking dimensions, allowing benefits from both the methods.


\begin{figure}[t]
    \centering
    \includegraphics[width=1.0\columnwidth]{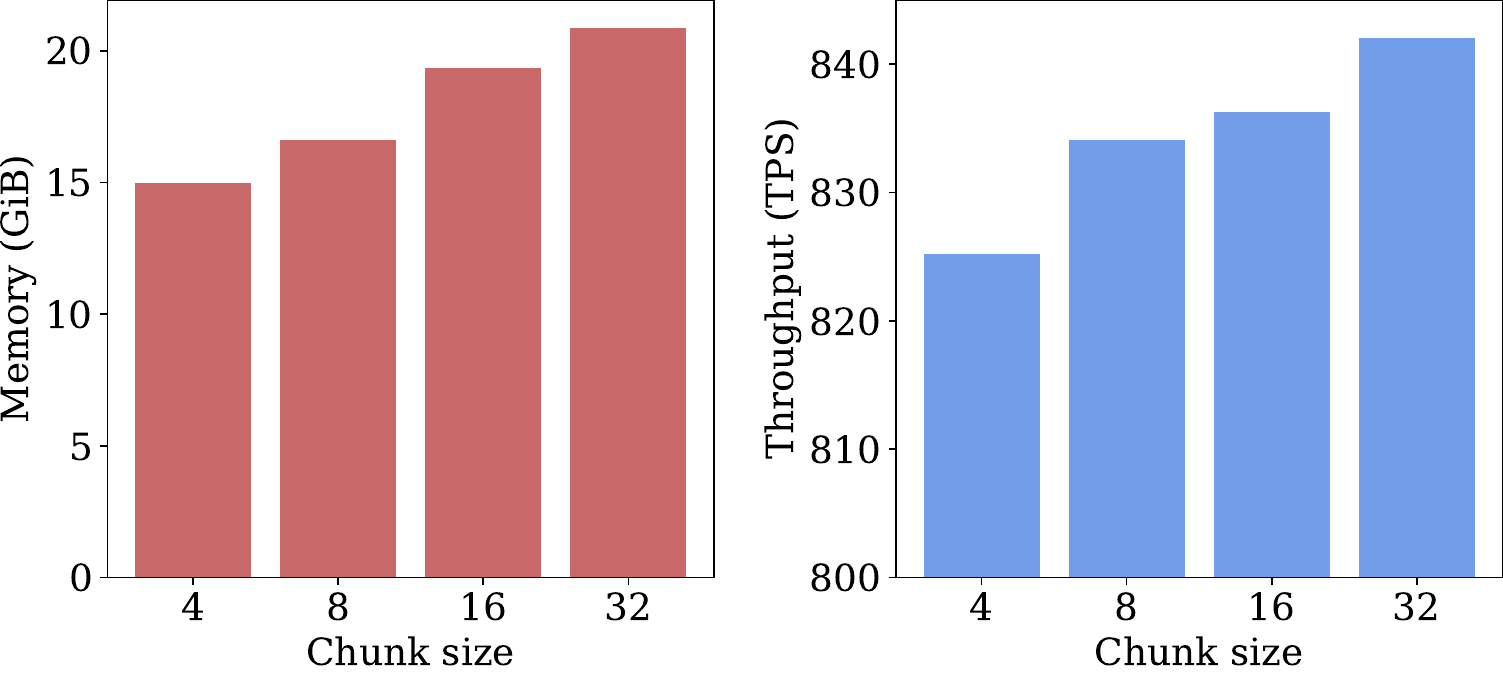}
    \caption{Ablation analysis of \ours{}'s head-chunk size $U$ when training Llama 3-8B with a context size of 512K on $C=$ 4 GPUs.}
    \vspace{-3.0ex}
    \label{fig:chunk_size}
\end{figure}


\subsection{Ablation on the head chunk size}
\label{sec:ablation}


For the above experiments, we aimed to show the maximum memory efficiency of \ours{}, so we chose $U$ = $C$. Next, we present an ablation on $U$, discussing the associated tradeoffs between memory and throughput.
Specifically, we analyze the memory-runtime tradeoff associated with the number of chunks and the size of each chunk. 
We use Llama 3-8B on 4$\times$H100 GPUs with a context size of 512K tokens.

As shown in \cref{fig:chunk_size}, increasing $U$ corresponds to more heads processed per stage, resulting in higher memory usage and lower runtime. 
Conversely, when $U$ = $C$, \ours{} provides maximum memory benefits at the cost of slight performance degradation due to kernel launch overhead. 
However, at longer context lengths, this overhead is amortized by the increased length, as shown in Table \ref{tab:runtime} of Appendix \ref{app:c}.

\section{Reinvesting Memory Savings}
UPipe consumes significantly less memory than Ulysses. 
However, for context lengths in the 128K--1M range, UPipe falls slightly behind Ulysses in terms of performance. 
In this section, we demonstrate the potential performance gains that could be obtained by reinvesting UPipe's memory savings.

\vspace{-1ex}
\subsection{Strided Activation Offloading}
As mentioned in Section \ref{subsec:setup}, we employ full activation checkpointing with CPU offloading. 
Since UPipe provides more GPU memory headroom, we explore \textbf{strided activation offloading}, which skips offloading the activations of every $K^{th}$ layer. 
This eliminates the CPU--GPU transfer overhead for every $K^{th}$ layer, resulting in an overall performance improvement.

\begin{table}[h]
    \centering
    \caption{Comparison of throughput (TPS) and memory usage (GiB) across Ulysses, UPipe, and UPipe with \textbf{SAO} (Strided Activation Offloading). \textbf{SAO} narrows the performance gap between Ulysses and UPipe due to less frequent activation offloading to CPU. }
    \begin{tabular*}{\columnwidth}{@{}rr@{\extracolsep{\fill}}ccccc}
    \toprule
    \vspace{1pt}
    & Method &\phantom{xx} 128K & 256K & 512K & 1M \\
    \midrule
    \multirow{3}{*}[0em]{\rotatebox[origin=c]{90}{TPS}}
         &Ulysses &\phantom{xx} 2417.79 & \textbf{1516.51} & \textbf{880.27} & \textbf{476.45} \\
         &UPipe &\phantom{xx} 2377.98 & 1515.59 & 878.73 & 470.34 \\
         &UPipe-SAO &\phantom{xx} \textbf{2445.10} & 1512.69 & 872.57 & 471.07 \\
    \midrule
    \multirow{3}{*}[0em]{\rotatebox[origin=c]{90}{GiB}}
         &Ulysses &\phantom{xx} 62.94 & 63.01 & 63.15 & 63.43 \\
         &UPipe &\phantom{xx} 53.24 & 53.31 & 53.45 & 53.73 \\
         &UPipe-SAO &\phantom{xx} 62.39 & 62.47 & 62.61 & 62.89 \\
    \bottomrule
    \end{tabular*}
    \label{tab:sao}
    \vspace{-1ex}
\end{table}

We conduct strided activation offloading experiments on Llama 3-8B using a single 8$\times$H100 node. 
We use a batch size $>1$ such that the total number of tokens is always 3M.
We use $K=7$, i.e., skip CPU activation offloading every 7 layers.
Table~\ref{tab:sao} compares throughput and memory usage across Ulysses, UPipe and UPipe-SAO.
For the context length of 128K, SAO significantly improves performance over both Ulysses and UPipe.
This is because at shorter sequence lengths, the CPU--GPU transfers contribute significantly to the overall runtime. 
Skipping CPU offloading for every $7^{th}$ layer results in \textbf{2.8\%} improvement in performance for 128K.
For longer sequence lengths, we see minimal change, because the quadratic scaling of attention computation time dominates the overall runtime.

\vspace{-1ex}
\subsection{Selective Activation Checkpointing}

\label{app:sac}

Selective activation checkpointing (SAC,~\citealp{megatron-sp}) retains the output activations of attention to prevent costly recomputation during the backward pass. 
The forward pass of UPipe is implemented as a custom PyTorch operator, enabling off-the-shelf support for SAC.
As shown in Table \ref{tab:sac}, enabling SAC further improves the performance of \ours{}, while keeping its memory usage lower than that of DeepSpeed-Ulysses (both with and without SAC).

\begin{table}[h]
    \centering
    \small
    \caption{Throughput (in tokens/second) and memory usage (GiB) of Ulysses and UPipe with and without Selective Activation Checkpointing at a sequence length of 512K for Llama 3-8B.}
    \begin{tabular}{lcc}
    \toprule
    Method & Throughput & Memory \\
    \midrule
    Ulysses & 864.16 & 27.34 \\
    Ulysses+SAC  & \textbf{1066.01} & 27.85 \\
    UPipe   & 850.97 & \textbf{25.24} \\
    UPipe+SAC    & 1056.77 & 25.75 \\
    \bottomrule
    \label{tab:sac}
    \end{tabular}
    \vspace{-5ex}
\end{table}

\subsection{Batch size ablation}

Due to the lower memory usage of \ours{}, it is also possible to reuse the freed up memory by increasing the per-device batch size.
Hence, we analyze the training throughput by varying the batch size for Llama3-8B training on a single 8$\times$H100 node at 128K context length.
We turn off CPU offloading of activations to avoid performance variability due to the PCIe bus traffic. 
As shown in Table \ref{tab:bs}, the throughput increases monotonically for batch sizes smaller than 8. 
However, at the batch size of 10, the speed of Ulysses degrades due to memory pressure triggering frequent CUDA allocation retries. 
UPipe avoids this through better memory management, outperforming Ulysses at the largest batch size while operating at lower memory across all batch sizes.

\begin{table}[h]
    \centering
    \small
    \vspace{-0.5ex}
    \caption{Comparison of throughput (TPS) and memory usage (GiB) for Ulysses and UPipe across different batch sizes.}
    \setlength{\tabcolsep}{4pt}
    \begin{tabular*}{\columnwidth}{@{}cr@{\extracolsep{\fill}}ccccc@{}}
    \toprule
    & Batch size\phantom{x} & 1 & 2 & 4 & 8 & 10 \\
    \midrule
    \multirow{2}{*}{\rotatebox[origin=c]{90}{TPS}}
         & Ulysses\phantom{x} & \textbf{2458.28} & \textbf{2485.78} & \textbf{2543.43} & \textbf{2564.38} & 2464.10 \\
         & UPipe\phantom{x}   & 2405.30 & 2432.11 & 2498.80 & 2518.56 & \textbf{2526.77} \\
    \midrule
    \multirow{2}{*}{\rotatebox[origin=c]{90}{GiB}}
         & Ulysses\phantom{x} & 22.53 & 27.25 & 38.13 & 59.99 & 71.13 \\
         & UPipe\phantom{x}   & \textbf{22.53} & \textbf{27.03} & \textbf{36.56} & \textbf{56.79} & \textbf{67.11} \\
    \bottomrule
    \end{tabular*}
    \label{tab:bs}
\end{table}


\vspace{-3ex}
\section{Conclusion}

In this paper, we introduced \ours{}, a context parallelism method that reduces the attention activation memory usage by chunking attention at the head level.
Our experiments demonstrate that these memory gains allow further context scaling without sacrificing throughput.
On Llama 3-8B, \ours{} reaches \textbf{5M }tokens on a single 8$\times$H100 node (\textbf{25\%} beyond FPDT), and scales to \textbf{8M} tokens on 16 H100 GPUs, while keeping throughput comparable to common baselines.
For larger models, \ours{} reduces the intermediate tensor memory usage of attention by up to 87.5\%, helping avoid allocation retries that otherwise degrade training performance.

Overall, \ours{} provides a practical path to pushing the context length frontier: it is simple, composable with existing methods, and offers substantial memory usage reductions while preserving the training speed.
While the study of \ours{} could be extended (see Appendix \ref{app:limitations} for an overview of limitations), we expect this method be a useful building block for future systems, especially as new tasks and modalities continue to demand larger sequence lengths.

\section*{Acknowledgements}

We thank the anonymous reviewers for their thoughtful and constructive feedback, which substantially improved this work. 
We are also grateful to the developers and maintainers of TorchTitan~\cite{torchtitan} and USP~\cite{usp}, whose open-source projects provided a strong basis for the implementation and evaluation of \ours{}. 
More broadly, we thank the open-source and research community, whose tooling and prior results made this work possible.

\section*{Impact Statement}
This work aims to advance the field of Machine Learning by improving the memory efficiency of training transformer based models on long context inputs. 
Our work extends the maximum context length that can fit on a given hardware, and thus is broadly applicable to a variety of long-context settings for Transformer models, but it is not designed for any particular use case.

\bibliography{bibliography}

@article{minimaxm1,
  title={MiniMax-M1: Scaling Test-Time Compute Efficiently with Lightning Attention},
  author={Chen, Aili and Li, Aonian and Gong, Bangwei and Jiang, Binyang and Fei, Bo and Yang, Bo and Shan, Boji and Yu, Changqing and Wang, Chao and Zhu, Cheng and others},
  journal={arXiv preprint arXiv:2506.13585},
  year={2025}
}

@article{gemini3,
  title = {Gemini 3.0: A New Era of Intelligence with Gemini 3},
  author = {{Gemini Team}},
  journal = {Google DeepMind},
  year = {2025},
  url = {https://blog.google/products-and-platforms/products/gemini/gemini-3/}
}

@misc{kimik2,
      title={Kimi K2: Open Agentic Intelligence}, 
      author={{Kimi Team} and Yifan Bai and Yiping Bao and Guanduo Chen and Jiahao Chen and Ningxin Chen and Ruijue Chen and Yanru Chen and Yuankun Chen and Yutian Chen and Zhuofu Chen and Jialei Cui and Hao Ding and Mengnan Dong and Angang Du and Chenzhuang Du and Dikang Du and Yulun Du and Yu Fan and Yichen Feng and Kelin Fu and Bofei Gao and Hongcheng Gao and Peizhong Gao and Tong Gao and Xinran Gu and Longyu Guan and Haiqing Guo and Jianhang Guo and Hao Hu and Xiaoru Hao and Tianhong He and Weiran He and Wenyang He and Chao Hong and Yangyang Hu and Zhenxing Hu and Weixiao Huang and Zhiqi Huang and Zihao Huang and Tao Jiang and Zhejun Jiang and Xinyi Jin and Yongsheng Kang and Guokun Lai and Cheng Li and Fang Li and Haoyang Li and Ming Li and Wentao Li and Yanhao Li and Yiwei Li and Zhaowei Li and Zheming Li and Hongzhan Lin and Xiaohan Lin and Zongyu Lin and Chengyin Liu and Chenyu Liu and Hongzhang Liu and Jingyuan Liu and Junqi Liu and Liang Liu and Shaowei Liu and T. Y. Liu and Tianwei Liu and Weizhou Liu and Yangyang Liu and Yibo Liu and Yiping Liu and Yue Liu and Zhengying Liu and Enzhe Lu and Lijun Lu and Shengling Ma and Xinyu Ma and Yingwei Ma and Shaoguang Mao and Jie Mei and Xin Men and Yibo Miao and Siyuan Pan and Yebo Peng and Ruoyu Qin and Bowen Qu and Zeyu Shang and Lidong Shi and Shengyuan Shi and Feifan Song and Jianlin Su and Zhengyuan Su and Xinjie Sun and Flood Sung and Heyi Tang and Jiawen Tao and Qifeng Teng and Chensi Wang and Dinglu Wang and Feng Wang and Haiming Wang and Jianzhou Wang and Jiaxing Wang and Jinhong Wang and Shengjie Wang and Shuyi Wang and Yao Wang and Yejie Wang and Yiqin Wang and Yuxin Wang and Yuzhi Wang and Zhaoji Wang and Zhengtao Wang and Zhexu Wang and Chu Wei and Qianqian Wei and Wenhao Wu and Xingzhe Wu and Yuxin Wu and Chenjun Xiao and Xiaotong Xie and Weimin Xiong and Boyu Xu and Jing Xu and Jinjing Xu and L. H. Xu and Lin Xu and Suting Xu and Weixin Xu and Xinran Xu and Yangchuan Xu and Ziyao Xu and Junjie Yan and Yuzi Yan and Xiaofei Yang and Ying Yang and Zhen Yang and Zhilin Yang and Zonghan Yang and Haotian Yao and Xingcheng Yao and Wenjie Ye and Zhuorui Ye and Bohong Yin and Longhui Yu and Enming Yuan and Hongbang Yuan and Mengjie Yuan and Haobing Zhan and Dehao Zhang and Hao Zhang and Wanlu Zhang and Xiaobin Zhang and Yangkun Zhang and Yizhi Zhang and Yongting Zhang and Yu Zhang and Yutao Zhang and Yutong Zhang and Zheng Zhang and Haotian Zhao and Yikai Zhao and Huabin Zheng and Shaojie Zheng and Jianren Zhou and Xinyu Zhou and Zaida Zhou and Zhen Zhu and Weiyu Zhuang and Xinxing Zu},
      year={2025},
      eprint={2507.20534},
      archivePrefix={arXiv},
      primaryClass={cs.LG},
      url={https://arxiv.org/abs/2507.20534}, 
}

@misc{magi1,
      title={MAGI-1: Autoregressive Video Generation at Scale}, 
      author={{Sand.ai} and Hansi Teng and Hongyu Jia and Lei Sun and Lingzhi Li and Maolin Li and Mingqiu Tang and Shuai Han and Tianning Zhang and W. Q. Zhang and Weifeng Luo and Xiaoyang Kang and Yuchen Sun and Yue Cao and Yunpeng Huang and Yutong Lin and Yuxin Fang and Zewei Tao and Zheng Zhang and Zhongshu Wang and Zixun Liu and Dai Shi and Guoli Su and Hanwen Sun and Hong Pan and Jie Wang and Jiexin Sheng and Min Cui and Min Hu and Ming Yan and Shucheng Yin and Siran Zhang and Tingting Liu and Xianping Yin and Xiaoyu Yang and Xin Song and Xuan Hu and Yankai Zhang and Yuqiao Li},
      year={2025},
      eprint={2505.13211},
      archivePrefix={arXiv},
      primaryClass={cs.CV},
      url={https://arxiv.org/abs/2505.13211}, 
}

@misc{wan,
      title={Wan: Open and Advanced Large-Scale Video Generative Models}, 
      author={{Team Wan} and Ang Wang and Baole Ai and Bin Wen and Chaojie Mao and Chen-Wei Xie and Di Chen and Feiwu Yu and Haiming Zhao and Jianxiao Yang and Jianyuan Zeng and Jiayu Wang and Jingfeng Zhang and Jingren Zhou and Jinkai Wang and Jixuan Chen and Kai Zhu and Kang Zhao and Keyu Yan and Lianghua Huang and Mengyang Feng and Ningyi Zhang and Pandeng Li and Pingyu Wu and Ruihang Chu and Ruili Feng and Shiwei Zhang and Siyang Sun and Tao Fang and Tianxing Wang and Tianyi Gui and Tingyu Weng and Tong Shen and Wei Lin and Wei Wang and Wei Wang and Wenmeng Zhou and Wente Wang and Wenting Shen and Wenyuan Yu and Xianzhong Shi and Xiaoming Huang and Xin Xu and Yan Kou and Yangyu Lv and Yifei Li and Yijing Liu and Yiming Wang and Yingya Zhang and Yitong Huang and Yong Li and You Wu and Yu Liu and Yulin Pan and Yun Zheng and Yuntao Hong and Yupeng Shi and Yutong Feng and Zeyinzi Jiang and Zhen Han and Zhi-Fan Wu and Ziyu Liu},
      year={2025},
      eprint={2503.20314},
      archivePrefix={arXiv},
      primaryClass={cs.CV},
      url={https://arxiv.org/abs/2503.20314}, 
}

@misc{hunyuanvideo,
      title={HunyuanVideo 1.5 Technical Report}, 
      author={Bing Wu and Chang Zou and Changlin Li and Duojun Huang and Fang Yang and Hao Tan and Jack Peng and Jianbing Wu and Jiangfeng Xiong and Jie Jiang and Linus and Patrol and Peizhen Zhang and Peng Chen and Penghao Zhao and Qi Tian and Songtao Liu and Weijie Kong and Weiyan Wang and Xiao He and Xin Li and Xinchi Deng and Xuefei Zhe and Yang Li and Yanxin Long and Yuanbo Peng and Yue Wu and Yuhong Liu and Zhenyu Wang and Zuozhuo Dai and Bo Peng and Coopers Li and Gu Gong and Guojian Xiao and Jiahe Tian and Jiaxin Lin and Jie Liu and Jihong Zhang and Jiesong Lian and Kaihang Pan and Lei Wang and Lin Niu and Mingtao Chen and Mingyang Chen and Mingzhe Zheng and Miles Yang and Qiangqiang Hu and Qi Yang and Qiuyong Xiao and Runzhou Wu and Ryan Xu and Rui Yuan and Shanshan Sang and Shisheng Huang and Siruis Gong and Shuo Huang and Weiting Guo and Xiang Yuan and Xiaojia Chen and Xiawei Hu and Wenzhi Sun and Xiele Wu and Xianshun Ren and Xiaoyan Yuan and Xiaoyue Mi and Yepeng Zhang and Yifu Sun and Yiting Lu and Yitong Li and You Huang and Yu Tang and Yixuan Li and Yuhang Deng and Yuan Zhou and Zhichao Hu and Zhiguang Liu and Zhihe Yang and Zilin Yang and Zhenzhi Lu and Zixiang Zhou and Zhao Zhong},
      year={2025},
      eprint={2511.18870},
      archivePrefix={arXiv},
      primaryClass={cs.CV},
      url={https://arxiv.org/abs/2511.18870}, 
}

@inproceedings{transformer,
  title     = {Attention Is All You Need},
  author    = {Vaswani, Ashish and Shazeer, Noam and Parmar, Niki and Uszkoreit, Jakob and Jones, Llion and Gomez, Aidan N. and Kaiser, Lukasz and Polosukhin, Illia},
  booktitle = {Advances in Neural Information Processing Systems (NeurIPS)},
  year      = {2017}
}

@misc{starcoder,
      title={StarCoder: may the source be with you!}, 
      author={Raymond Li and Loubna Ben Allal and Yangtian Zi and Niklas Muennighoff and Denis Kocetkov and Chenghao Mou and Marc Marone and Christopher Akiki and Jia Li and Jenny Chim and Qian Liu and Evgenii Zheltonozhskii and Terry Yue Zhuo and Thomas Wang and Olivier Dehaene and Mishig Davaadorj and Joel Lamy-Poirier and João Monteiro and Oleh Shliazhko and Nicolas Gontier and Nicholas Meade and Armel Zebaze and Ming-Ho Yee and Logesh Kumar Umapathi and Jian Zhu and Benjamin Lipkin and Muhtasham Oblokulov and Zhiruo Wang and Rudra Murthy and Jason Stillerman and Siva Sankalp Patel and Dmitry Abulkhanov and Marco Zocca and Manan Dey and Zhihan Zhang and Nour Fahmy and Urvashi Bhattacharyya and Wenhao Yu and Swayam Singh and Sasha Luccioni and Paulo Villegas and Maxim Kunakov and Fedor Zhdanov and Manuel Romero and Tony Lee and Nadav Timor and Jennifer Ding and Claire Schlesinger and Hailey Schoelkopf and Jan Ebert and Tri Dao and Mayank Mishra and Alex Gu and Jennifer Robinson and Carolyn Jane Anderson and Brendan Dolan-Gavitt and Danish Contractor and Siva Reddy and Daniel Fried and Dzmitry Bahdanau and Yacine Jernite and Carlos Muñoz Ferrandis and Sean Hughes and Thomas Wolf and Arjun Guha and Leandro von Werra and Harm de Vries},
      year={2023},
      eprint={2305.06161},
      archivePrefix={arXiv},
      primaryClass={cs.CL},
      url={https://arxiv.org/abs/2305.06161}, 
}

@misc{qwencoder,
      title={Qwen2.5-Coder Technical Report}, 
      author={Binyuan Hui and Jian Yang and Zeyu Cui and Jiaxi Yang and Dayiheng Liu and Lei Zhang and Tianyu Liu and Jiajun Zhang and Bowen Yu and Keming Lu and Kai Dang and Yang Fan and Yichang Zhang and An Yang and Rui Men and Fei Huang and Bo Zheng and Yibo Miao and Shanghaoran Quan and Yunlong Feng and Xingzhang Ren and Xuancheng Ren and Jingren Zhou and Junyang Lin},
      year={2024},
      eprint={2409.12186},
      archivePrefix={arXiv},
      primaryClass={cs.CL},
      url={https://arxiv.org/abs/2409.12186}, 
}

@misc{mlongdoc,
      title={M-Longdoc: A Benchmark For Multimodal Super-Long Document Understanding And A Retrieval-Aware Tuning Framework}, 
      author={Yew Ken Chia and Liying Cheng and Hou Pong Chan and Chaoqun Liu and Maojia Song and Sharifah Mahani Aljunied and Soujanya Poria and Lidong Bing},
      year={2024},
      eprint={2411.06176},
      archivePrefix={arXiv},
      primaryClass={cs.CL},
      url={https://arxiv.org/abs/2411.06176}, 
}

@misc{longrag,
      title={LongRAG: Enhancing Retrieval-Augmented Generation with Long-context LLMs}, 
      author={Ziyan Jiang and Xueguang Ma and Wenhu Chen},
      year={2024},
      eprint={2406.15319},
      archivePrefix={arXiv},
      primaryClass={cs.CL},
      url={https://arxiv.org/abs/2406.15319}, 
}

@article{audio,
  title={Advanced Long-context End-to-end Speech Recognition Using Context-expanded Transformers},
  author={Takaaki Hori and Niko Moritz and Chiori Hori and Jonathan Le Roux},
  journal={ArXiv},
  year={2021},
  volume={abs/2104.09426},
  url={https://api.semanticscholar.org/CorpusID:233296591}
}

@misc{seqparallel,
      title={Sequence Parallelism: Long Sequence Training from System Perspective}, 
      author={Shenggui Li and Fuzhao Xue and Chaitanya Baranwal and Yongbin Li and Yang You},
      year={2022},
      eprint={2105.13120},
      archivePrefix={arXiv},
      primaryClass={cs.LG},
      url={https://arxiv.org/abs/2105.13120}, 
}

@misc{ring_new,
      title={Ring Attention with Blockwise Transformers for Near-Infinite Context}, 
      author={Hao Liu and Matei Zaharia and Pieter Abbeel},
      year={2023},
      eprint={2310.01889},
      archivePrefix={arXiv},
      primaryClass={cs.CL},
      url={https://arxiv.org/abs/2310.01889}, 
}

@inproceedings{ring,
    title = "Sequence Parallelism: Long Sequence Training from System Perspective",
    author = "Li, Shenggui  and
      Xue, Fuzhao  and
      Baranwal, Chaitanya  and
      Li, Yongbin  and
      You, Yang",
    editor = "Rogers, Anna  and
      Boyd-Graber, Jordan  and
      Okazaki, Naoaki",
    booktitle = "Proceedings of the 61st Annual Meeting of the Association for Computational Linguistics (Volume 1: Long Papers)",
    month = jul,
    year = "2023",
    address = "Toronto, Canada",
    publisher = "Association for Computational Linguistics",
    url = "https://aclanthology.org/2023.acl-long.134/",
    doi = "10.18653/v1/2023.acl-long.134",
    pages = "2391--2404"
}

@misc{ulysses,
      title={DeepSpeed Ulysses: System Optimizations for Enabling Training of Extreme Long Sequence Transformer Models}, 
      author={Sam Ade Jacobs and Masahiro Tanaka and Chengming Zhang and Minjia Zhang and Shuaiwen Leon Song and Samyam Rajbhandari and Yuxiong He},
      year={2023},
      eprint={2309.14509},
      archivePrefix={arXiv},
      primaryClass={cs.LG},
      url={https://arxiv.org/abs/2309.14509}, 
}

@misc{usp,
      title={USP: A Unified Sequence Parallelism Approach for Long Context Generative AI}, 
      author={Jiarui Fang and Shangchun Zhao},
      year={2024},
      eprint={2405.07719},
      archivePrefix={arXiv},
      primaryClass={cs.LG},
      url={https://arxiv.org/abs/2405.07719}, 
}

@misc{gqa,
      title={GQA: Training Generalized Multi-Query Transformer Models from Multi-Head Checkpoints}, 
      author={Joshua Ainslie and James Lee-Thorp and Michiel de Jong and Yury Zemlyanskiy and Federico Lebrón and Sumit Sanghai},
      year={2023},
      eprint={2305.13245},
      archivePrefix={arXiv},
      primaryClass={cs.CL},
      url={https://arxiv.org/abs/2305.13245}, 
}

@article{speech,
  title={SimpleSpeech: Towards Simple and Efficient Text-to-Speech with Scalar Latent Transformer Diffusion Models},
  author={Dongchao Yang and Dingdong Wang and Haohan Guo and Xueyuan Chen and Xixin Wu and Helen M. Meng},
  journal={ArXiv},
  year={2024},
  volume={abs/2406.02328},
  url={https://api.semanticscholar.org/CorpusID:270226637}
}

@misc{alst,
      title={Arctic Long Sequence Training: Scalable And Efficient Training For Multi-Million Token Sequences}, 
      author={Stas Bekman and Samyam Rajbhandari and Michael Wyatt and Jeff Rasley and Tunji Ruwase and Zhewei Yao and Aurick Qiao and Yuxiong He},
      year={2025},
      eprint={2506.13996},
      archivePrefix={arXiv},
      primaryClass={cs.LG},
      url={https://arxiv.org/abs/2506.13996}, 
}

@misc{fpdt,
      title={Training Ultra Long Context Language Model with Fully Pipelined Distributed Transformer}, 
      author={Jinghan Yao and Sam Ade Jacobs and Masahiro Tanaka and Olatunji Ruwase and Hari Subramoni and Dhabaleswar K. Panda},
      year={2025},
      eprint={2408.16978},
      archivePrefix={arXiv},
      primaryClass={cs.DC},
      url={https://arxiv.org/abs/2408.16978}, 
}

@inproceedings{
liger,
title={Liger-Kernel: Efficient Triton Kernels for {LLM} Training},
author={Pin-Lun Hsu and Yun Dai and Vignesh Kothapalli and Qingquan Song and Shao Tang and Siyu Zhu and Steven Shimizu and Shivam Sahni and Haowen Ning and Yanning Chen and Zhipeng Wang},
booktitle={Championing Open-source DEvelopment in ML Workshop @ ICML25},
year={2025},
url={https://openreview.net/forum?id=36SjAIT42G}
}

@inproceedings{
   torchtitan,
   title={TorchTitan: One-stop PyTorch native solution for production ready {LLM} pretraining},
   author={Wanchao Liang and Tianyu Liu and Less Wright and Will Constable and Andrew Gu and Chien-Chin Huang and Iris Zhang and Wei Feng and Howard Huang and Junjie Wang and Sanket Purandare and Gokul Nadathur and Stratos Idreos},
   booktitle={The Thirteenth International Conference on Learning Representations},
   year={2025},
   url={https://openreview.net/forum?id=SFN6Wm7YBI}
}

@misc{llama3,
      title={The Llama 3 Herd of Models}, 
      author={Aaron Grattafiori and Abhimanyu Dubey and Abhinav Jauhri and Abhinav Pandey and Abhishek Kadian and Ahmad Al-Dahle and Aiesha Letman and Akhil Mathur and Alan Schelten and Alex Vaughan and Amy Yang and Angela Fan and Anirudh Goyal and Anthony Hartshorn and Aobo Yang and Archi Mitra and Archie Sravankumar and Artem Korenev and Arthur Hinsvark and Arun Rao and Aston Zhang and Aurelien Rodriguez and Austen Gregerson and Ava Spataru and Baptiste Roziere and Bethany Biron and Binh Tang and Bobbie Chern and Charlotte Caucheteux and Chaya Nayak and Chloe Bi and Chris Marra and Chris McConnell and Christian Keller and Christophe Touret and Chunyang Wu and Corinne Wong and Cristian Canton Ferrer and Cyrus Nikolaidis and Damien Allonsius and Daniel Song and Danielle Pintz and Danny Livshits and Danny Wyatt and David Esiobu and Dhruv Choudhary and Dhruv Mahajan and Diego Garcia-Olano and Diego Perino and Dieuwke Hupkes and Egor Lakomkin and Ehab AlBadawy and Elina Lobanova and Emily Dinan and Eric Michael Smith and Filip Radenovic and Francisco Guzmán and Frank Zhang and Gabriel Synnaeve and Gabrielle Lee and Georgia Lewis Anderson and Govind Thattai and Graeme Nail and Gregoire Mialon and Guan Pang and Guillem Cucurell and Hailey Nguyen and Hannah Korevaar and Hu Xu and Hugo Touvron and Iliyan Zarov and Imanol Arrieta Ibarra and Isabel Kloumann and Ishan Misra and Ivan Evtimov and Jack Zhang and Jade Copet and Jaewon Lee and Jan Geffert and Jana Vranes and Jason Park and Jay Mahadeokar and Jeet Shah and Jelmer van der Linde and Jennifer Billock and Jenny Hong and Jenya Lee and Jeremy Fu and Jianfeng Chi and Jianyu Huang and Jiawen Liu and Jie Wang and Jiecao Yu and Joanna Bitton and Joe Spisak and Jongsoo Park and Joseph Rocca and Joshua Johnstun and Joshua Saxe and Junteng Jia and Kalyan Vasuden Alwala and Karthik Prasad and Kartikeya Upasani and Kate Plawiak and Ke Li and Kenneth Heafield and Kevin Stone and Khalid El-Arini and Krithika Iyer and Kshitiz Malik and Kuenley Chiu and Kunal Bhalla and Kushal Lakhotia and Lauren Rantala-Yeary and Laurens van der Maaten and Lawrence Chen and Liang Tan and Liz Jenkins and Louis Martin and Lovish Madaan and Lubo Malo and Lukas Blecher and Lukas Landzaat and Luke de Oliveira and Madeline Muzzi and Mahesh Pasupuleti and Mannat Singh and Manohar Paluri and Marcin Kardas and Maria Tsimpoukelli and Mathew Oldham and Mathieu Rita and Maya Pavlova and Melanie Kambadur and Mike Lewis and Min Si and Mitesh Kumar Singh and Mona Hassan and Naman Goyal and Narjes Torabi and Nikolay Bashlykov and Nikolay Bogoychev and Niladri Chatterji and Ning Zhang and Olivier Duchenne and Onur Çelebi and Patrick Alrassy and Pengchuan Zhang and Pengwei Li and Petar Vasic and Peter Weng and Prajjwal Bhargava and Pratik Dubal and Praveen Krishnan and Punit Singh Koura and Puxin Xu and Qing He and Qingxiao Dong and Ragavan Srinivasan and Raj Ganapathy and Ramon Calderer and Ricardo Silveira Cabral and Robert Stojnic and Roberta Raileanu and Rohan Maheswari and Rohit Girdhar and Rohit Patel and Romain Sauvestre and Ronnie Polidoro and Roshan Sumbaly and Ross Taylor and Ruan Silva and Rui Hou and Rui Wang and Saghar Hosseini and Sahana Chennabasappa and Sanjay Singh and Sean Bell and Seohyun Sonia Kim and Sergey Edunov and Shaoliang Nie and Sharan Narang and Sharath Raparthy and Sheng Shen and Shengye Wan and Shruti Bhosale and Shun Zhang and Simon Vandenhende and Soumya Batra and Spencer Whitman and Sten Sootla and Stephane Collot and Suchin Gururangan and Sydney Borodinsky and Tamar Herman and Tara Fowler and Tarek Sheasha and Thomas Georgiou and Thomas Scialom and Tobias Speckbacher and Todor Mihaylov and Tong Xiao and Ujjwal Karn and Vedanuj Goswami and Vibhor Gupta and Vignesh Ramanathan and Viktor Kerkez and Vincent Gonguet and Virginie Do and Vish Vogeti and Vítor Albiero and Vladan Petrovic and Weiwei Chu and Wenhan Xiong and Wenyin Fu and Whitney Meers and Xavier Martinet and Xiaodong Wang and Xiaofang Wang and Xiaoqing Ellen Tan and Xide Xia and Xinfeng Xie and Xuchao Jia and Xuewei Wang and Yaelle Goldschlag and Yashesh Gaur and Yasmine Babaei and Yi Wen and Yiwen Song and Yuchen Zhang and Yue Li and Yuning Mao and Zacharie Delpierre Coudert and Zheng Yan and Zhengxing Chen and Zoe Papakipos and Aaditya Singh and Aayushi Srivastava and Abha Jain and Adam Kelsey and Adam Shajnfeld and Adithya Gangidi and Adolfo Victoria and Ahuva Goldstand and Ajay Menon and Ajay Sharma and Alex Boesenberg and Alexei Baevski and Allie Feinstein and Amanda Kallet and Amit Sangani and Amos Teo and Anam Yunus and Andrei Lupu and Andres Alvarado and Andrew Caples and Andrew Gu and Andrew Ho and Andrew Poulton and Andrew Ryan and Ankit Ramchandani and Annie Dong and Annie Franco and Anuj Goyal and Aparajita Saraf and Arkabandhu Chowdhury and Ashley Gabriel and Ashwin Bharambe and Assaf Eisenman and Azadeh Yazdan and Beau James and Ben Maurer and Benjamin Leonhardi and Bernie Huang and Beth Loyd and Beto De Paola and Bhargavi Paranjape and Bing Liu and Bo Wu and Boyu Ni and Braden Hancock and Bram Wasti and Brandon Spence and Brani Stojkovic and Brian Gamido and Britt Montalvo and Carl Parker and Carly Burton and Catalina Mejia and Ce Liu and Changhan Wang and Changkyu Kim and Chao Zhou and Chester Hu and Ching-Hsiang Chu and Chris Cai and Chris Tindal and Christoph Feichtenhofer and Cynthia Gao and Damon Civin and Dana Beaty and Daniel Kreymer and Daniel Li and David Adkins and David Xu and Davide Testuggine and Delia David and Devi Parikh and Diana Liskovich and Didem Foss and Dingkang Wang and Duc Le and Dustin Holland and Edward Dowling and Eissa Jamil and Elaine Montgomery and Eleonora Presani and Emily Hahn and Emily Wood and Eric-Tuan Le and Erik Brinkman and Esteban Arcaute and Evan Dunbar and Evan Smothers and Fei Sun and Felix Kreuk and Feng Tian and Filippos Kokkinos and Firat Ozgenel and Francesco Caggioni and Frank Kanayet and Frank Seide and Gabriela Medina Florez and Gabriella Schwarz and Gada Badeer and Georgia Swee and Gil Halpern and Grant Herman and Grigory Sizov and Guangyi and Zhang and Guna Lakshminarayanan and Hakan Inan and Hamid Shojanazeri and Han Zou and Hannah Wang and Hanwen Zha and Haroun Habeeb and Harrison Rudolph and Helen Suk and Henry Aspegren and Hunter Goldman and Hongyuan Zhan and Ibrahim Damlaj and Igor Molybog and Igor Tufanov and Ilias Leontiadis and Irina-Elena Veliche and Itai Gat and Jake Weissman and James Geboski and James Kohli and Janice Lam and Japhet Asher and Jean-Baptiste Gaya and Jeff Marcus and Jeff Tang and Jennifer Chan and Jenny Zhen and Jeremy Reizenstein and Jeremy Teboul and Jessica Zhong and Jian Jin and Jingyi Yang and Joe Cummings and Jon Carvill and Jon Shepard and Jonathan McPhie and Jonathan Torres and Josh Ginsburg and Junjie Wang and Kai Wu and Kam Hou U and Karan Saxena and Kartikay Khandelwal and Katayoun Zand and Kathy Matosich and Kaushik Veeraraghavan and Kelly Michelena and Keqian Li and Kiran Jagadeesh and Kun Huang and Kunal Chawla and Kyle Huang and Lailin Chen and Lakshya Garg and Lavender A and Leandro Silva and Lee Bell and Lei Zhang and Liangpeng Guo and Licheng Yu and Liron Moshkovich and Luca Wehrstedt and Madian Khabsa and Manav Avalani and Manish Bhatt and Martynas Mankus and Matan Hasson and Matthew Lennie and Matthias Reso and Maxim Groshev and Maxim Naumov and Maya Lathi and Meghan Keneally and Miao Liu and Michael L. Seltzer and Michal Valko and Michelle Restrepo and Mihir Patel and Mik Vyatskov and Mikayel Samvelyan and Mike Clark and Mike Macey and Mike Wang and Miquel Jubert Hermoso and Mo Metanat and Mohammad Rastegari and Munish Bansal and Nandhini Santhanam and Natascha Parks and Natasha White and Navyata Bawa and Nayan Singhal and Nick Egebo and Nicolas Usunier and Nikhil Mehta and Nikolay Pavlovich Laptev and Ning Dong and Norman Cheng and Oleg Chernoguz and Olivia Hart and Omkar Salpekar and Ozlem Kalinli and Parkin Kent and Parth Parekh and Paul Saab and Pavan Balaji and Pedro Rittner and Philip Bontrager and Pierre Roux and Piotr Dollar and Polina Zvyagina and Prashant Ratanchandani and Pritish Yuvraj and Qian Liang and Rachad Alao and Rachel Rodriguez and Rafi Ayub and Raghotham Murthy and Raghu Nayani and Rahul Mitra and Rangaprabhu Parthasarathy and Raymond Li and Rebekkah Hogan and Robin Battey and Rocky Wang and Russ Howes and Ruty Rinott and Sachin Mehta and Sachin Siby and Sai Jayesh Bondu and Samyak Datta and Sara Chugh and Sara Hunt and Sargun Dhillon and Sasha Sidorov and Satadru Pan and Saurabh Mahajan and Saurabh Verma and Seiji Yamamoto and Sharadh Ramaswamy and Shaun Lindsay and Shaun Lindsay and Sheng Feng and Shenghao Lin and Shengxin Cindy Zha and Shishir Patil and Shiva Shankar and Shuqiang Zhang and Shuqiang Zhang and Sinong Wang and Sneha Agarwal and Soji Sajuyigbe and Soumith Chintala and Stephanie Max and Stephen Chen and Steve Kehoe and Steve Satterfield and Sudarshan Govindaprasad and Sumit Gupta and Summer Deng and Sungmin Cho and Sunny Virk and Suraj Subramanian and Sy Choudhury and Sydney Goldman and Tal Remez and Tamar Glaser and Tamara Best and Thilo Koehler and Thomas Robinson and Tianhe Li and Tianjun Zhang and Tim Matthews and Timothy Chou and Tzook Shaked and Varun Vontimitta and Victoria Ajayi and Victoria Montanez and Vijai Mohan and Vinay Satish Kumar and Vishal Mangla and Vlad Ionescu and Vlad Poenaru and Vlad Tiberiu Mihailescu and Vladimir Ivanov and Wei Li and Wenchen Wang and Wenwen Jiang and Wes Bouaziz and Will Constable and Xiaocheng Tang and Xiaojian Wu and Xiaolan Wang and Xilun Wu and Xinbo Gao and Yaniv Kleinman and Yanjun Chen and Ye Hu and Ye Jia and Ye Qi and Yenda Li and Yilin Zhang and Ying Zhang and Yossi Adi and Youngjin Nam and Yu and Wang and Yu Zhao and Yuchen Hao and Yundi Qian and Yunlu Li and Yuzi He and Zach Rait and Zachary DeVito and Zef Rosnbrick and Zhaoduo Wen and Zhenyu Yang and Zhiwei Zhao and Zhiyu Ma},
      year={2024},
      eprint={2407.21783},
      archivePrefix={arXiv},
      primaryClass={cs.AI},
      url={https://arxiv.org/abs/2407.21783}, 
}

@misc{mistral,
      title={Mistral 7B}, 
      author={Albert Q. Jiang and Alexandre Sablayrolles and Arthur Mensch and Chris Bamford and Devendra Singh Chaplot and Diego de las Casas and Florian Bressand and Gianna Lengyel and Guillaume Lample and Lucile Saulnier and Lélio Renard Lavaud and Marie-Anne Lachaux and Pierre Stock and Teven Le Scao and Thibaut Lavril and Thomas Wang and Timothée Lacroix and William El Sayed},
      year={2023},
      eprint={2310.06825},
      archivePrefix={arXiv},
      primaryClass={cs.CL},
      url={https://arxiv.org/abs/2310.06825}, 
}

@misc{qwen3,
      title={Qwen3 Technical Report}, 
      author={An Yang and Anfeng Li and Baosong Yang and Beichen Zhang and Binyuan Hui and Bo Zheng and Bowen Yu and Chang Gao and Chengen Huang and Chenxu Lv and Chujie Zheng and Dayiheng Liu and Fan Zhou and Fei Huang and Feng Hu and Hao Ge and Haoran Wei and Huan Lin and Jialong Tang and Jian Yang and Jianhong Tu and Jianwei Zhang and Jianxin Yang and Jiaxi Yang and Jing Zhou and Jingren Zhou and Junyang Lin and Kai Dang and Keqin Bao and Kexin Yang and Le Yu and Lianghao Deng and Mei Li and Mingfeng Xue and Mingze Li and Pei Zhang and Peng Wang and Qin Zhu and Rui Men and Ruize Gao and Shixuan Liu and Shuang Luo and Tianhao Li and Tianyi Tang and Wenbiao Yin and Xingzhang Ren and Xinyu Wang and Xinyu Zhang and Xuancheng Ren and Yang Fan and Yang Su and Yichang Zhang and Yinger Zhang and Yu Wan and Yuqiong Liu and Zekun Wang and Zeyu Cui and Zhenru Zhang and Zhipeng Zhou and Zihan Qiu},
      year={2025},
      eprint={2505.09388},
      archivePrefix={arXiv},
      primaryClass={cs.CL},
      url={https://arxiv.org/abs/2505.09388}, 
}

@misc{fa3,
      title={FlashAttention-3: Fast and Accurate Attention with Asynchrony and Low-precision}, 
      author={Jay Shah and Ganesh Bikshandi and Ying Zhang and Vijay Thakkar and Pradeep Ramani and Tri Dao},
      year={2024},
      eprint={2407.08608},
      archivePrefix={arXiv},
      primaryClass={cs.LG},
      url={https://arxiv.org/abs/2407.08608}, 
}

@misc{rope,
      title={RoFormer: Enhanced Transformer with Rotary Position Embedding}, 
      author={Jianlin Su and Yu Lu and Shengfeng Pan and Ahmed Murtadha and Bo Wen and Yunfeng Liu},
      year={2023},
      eprint={2104.09864},
      archivePrefix={arXiv},
      primaryClass={cs.CL},
      url={https://arxiv.org/abs/2104.09864}, 
}

@misc{unsloth,
  author = {Daniel Han and Michael Han and {Unsloth team}},
  title = {Unsloth},
  url = {https://github.com/unslothai/unsloth},
  year = {2023}
}

@inproceedings{flashattn,
  title={Flash{A}ttention: Fast and Memory-Efficient Exact Attention with {IO}-Awareness},
  author={Dao, Tri and Fu, Daniel Y. and Ermon, Stefano and Rudra, Atri and R{\'e}, Christopher},
  booktitle={Advances in Neural Information Processing Systems (NeurIPS)},
  year={2022}
}

@misc{megatron-sp,
      title={Reducing Activation Recomputation in Large Transformer Models}, 
      author={Vijay Korthikanti and Jared Casper and Sangkug Lym and Lawrence McAfee and Michael Andersch and Mohammad Shoeybi and Bryan Catanzaro},
      year={2022},
      eprint={2205.05198},
      archivePrefix={arXiv},
      primaryClass={cs.LG},
      url={https://arxiv.org/abs/2205.05198}, 
}

@misc{swiglu,
      title={GLU Variants Improve Transformer}, 
      author={Noam Shazeer},
      year={2020},
      eprint={2002.05202},
      archivePrefix={arXiv},
      primaryClass={cs.LG},
      url={https://arxiv.org/abs/2002.05202}, 
}

@misc{c4,
      title={Exploring the Limits of Transfer Learning with a Unified Text-to-Text Transformer}, 
      author={Colin Raffel and Noam Shazeer and Adam Roberts and Katherine Lee and Sharan Narang and Michael Matena and Yanqi Zhou and Wei Li and Peter J. Liu},
      year={2023},
      eprint={1910.10683},
      archivePrefix={arXiv},
      primaryClass={cs.LG},
      url={https://arxiv.org/abs/1910.10683}, 
}

@misc{loongtrain,
      title={LoongTrain: Efficient Training of Long-Sequence LLMs with Head-Context Parallelism}, 
      author={Diandian Gu and Peng Sun and Qinghao Hu and Ting Huang and Xun Chen and Yingtong Xiong and Guoteng Wang and Qiaoling Chen and Shangchun Zhao and Jiarui Fang and Yonggang Wen and Tianwei Zhang and Xin Jin and Xuanzhe Liu},
      year={2024},
      eprint={2406.18485},
      archivePrefix={arXiv},
      primaryClass={cs.DC},
      url={https://arxiv.org/abs/2406.18485}, 
}
\bibliographystyle{icml2026}

\newpage
\appendix
\onecolumn
\section*{Supplementary Material}

\section{Per-stage memory accounting for the attention block}
\label{app:a}

In Tables~\ref{tab:gqa_fwd_mem} and~\ref{tab:gqa_bwd_mem} we list the peak activation memory of the attention block under GQA at each stage, for DeepSpeed-Ulysses with and without activation offloading, FPDT, and \ours{}.
This section walks through how the entries for the Ulysses and \ours{} rows are obtained.

\begin{table}[h]
    \centering
    \caption{Peak activation memory during the backward pass of the attention stage under GQA, expressed in units of one local hidden-state shard ($\frac{S}{C}\cdot d_{model}$ \texttt{bf16} elements, i.e., $2\cdot\frac{S}{C}\cdot d_{model}$ bytes). $\pi$ represents the number of chunks in FPDT, $\nu$ represents the number of chunks in UPipe.}
    \label{tab:gqa_bwd_mem}
    \renewcommand{\arraystretch}{1.0}
    \resizebox{1.0\textwidth}{!}{%
    \begin{tabular}{@{}lcccc@{}}
        \toprule
        \textbf{Method} & \textbf{Before Bwd Attn} & \textbf{During out\_all\_to\_all} & \textbf{During Bwd Attn Kernel} & \textbf{During inp\_all\_to\_all} \\
        \midrule
        \textbf{Ulysses} & $L+2$ & $L+3$ & $L + 2\gamma + 2$ & $L + \gamma + 1$ \\
        \midrule
        \textbf{Ulysses + offloading} & $3$ & $4$ & $2\gamma + 3$ & $\gamma + 2$ \\
        \midrule
        \textbf{FPDT} & $\frac{1}{\pi}$ & $\frac{3}{\pi}$ & $\frac{2\gamma + 4}{\pi}$ & $\frac{\gamma + 2}{\pi}$ \\
        \midrule
        \textbf{Untied Ulysses} & $3$ & $3 + \frac{1}{\nu}$ & $\max\!\left(3 + \frac{2\gamma}{\nu},\; 2 + \frac{2\gamma}{\nu} + \frac{2(\nu - 1)}{\nu}\right)$ & $2 + \frac{\gamma}{\nu} + \frac{2(\nu - 1)}{\nu}$ \\
        \bottomrule
    \end{tabular}%
    }
\end{table}

\paragraph{DeepSpeed-Ulysses (forward).}
On entry to the attention block, only the saved layer input $X$ is live on GPU.
During \texttt{inp\_all\_to\_all}, the projected $Q, K, V$ ($\gamma$ units) and one head-distributed destination buffer ($1$ unit) are alive in addition to $X$ ($1$ unit).
Because the non-QKVPacked all-to-all communicates $Q, K, V$ sequentially, only one destination buffer is in flight at a time, and the peak (for $G > 1$) occurs during $Q$'s exchange: $1\,(X) + \gamma\,(Q,K,V) + 1\,(Q') = \gamma + 2$.
During the attention kernel, the live set is $\{X, Q', K', V', O'\}$, again summing to $\gamma + 2$.
During \texttt{out\_all\_to\_all}, the head-distributed buffers are released and only $\{X, O', O\}$ remain ($3$).
Without activation offloading, the inputs of the preceding $L-1$ layers are also held on GPU, which we account for by the leading $L$ term of the Ulysses row.

\paragraph{DeepSpeed-Ulysses (backward).}
The block begins with three sequence-sharded $1$-unit tensors: $X$ (the layer input, saved by the framework for the QKV-projection weight gradient), $O$ (the saved attention block output, kept in sequence-sharded layout because it is the per-token input to proj\_o and thus required for the proj\_o weight gradient), and the upstream gradient $\delta_O$ on $O$, summing to $3$.
The FlashAttention backward kernel, however, needs both quantities in head-distributed form, so two all-to-alls are issued during \texttt{out\_all\_to\_all}: one on $O$ (producing $O'$) and one on $\delta_O$ (producing $\delta_{O}'$).
Running these sequentially with slot reuse---the buffer of the first all-to-all's input is freed after the call and reused as the destination of the second---caps the peak at $4$: $X$, one in-flight source, its destination, and the still-alive second source.
The FlashAttention backward kernel then requires $\{Q', K', V', O', \delta_{O}', \delta Q', \delta K', \delta V'\}$ simultaneously (a $(2\gamma+2)$-unit footprint) on top of $X$ (the seq-sharded $O$ and $\delta_O$ have been consumed in the previous step), yielding $2\gamma + 3$.
The inverse \texttt{inp\_all\_to\_all} mirrors the forward and peaks at $\gamma + 2$.

\paragraph{Untied Ulysses (forward).}
\ours{} executes the attention block in $\nu = H/U$ stages, each processing $U$ heads.
A full-sized output buffer \texttt{final\_out} ($1$ unit) is pre-allocated at the start of the block.
Because \texttt{final\_out} is initially empty, slot $i$ of size $1/\nu$ can be transiently repurposed during stage $i$ to hold the head-distributed query $Q^{\prime i}$, eliminating a separate destination buffer for that all-to-all.
At \texttt{inp\_all\_to\_all} of stage $i$, the live tensors are $X$, \texttt{final\_out} (with slot $i$ doubling as $Q^{\prime i}$'s buffer), and the chunked $Q^i, K^i, V^i$ ($\gamma/\nu$), giving $2 + \gamma/\nu$.
At the attention kernel, $Q^{\prime i}$ still resides in slot $i$ of \texttt{final\_out}, $K^{\prime i}, V^{\prime i}$ ($2/(G\nu)$) and the head-distributed output $O^{\prime i}$ ($1/\nu$) coexist separately, summing again to $2 + \gamma/\nu$.
At \texttt{out\_all\_to\_all}, $O^{\prime i}$ is gathered into slot $i$ of \texttt{final\_out}; the only transient cost is one in-flight chunk, yielding $2 + 1/\nu$.
Across the whole stage, the intermediate buffers (chunked $Q, K, V$, all-to-all in/out, attention output) are $\nu$ times smaller than their Ulysses counterparts, at the price of two persistent $1$-unit tensors ($X$ and \texttt{final\_out}) instead of one.

\paragraph{Untied Ulysses (backward).}
The saved \texttt{final\_out} and its upstream gradient $\delta_{\texttt{final\_out}}$ each remain alive throughout the $\nu$-stage loop, contributing $2$ units on top of $X$ on entry to the block.
Each stage consumes one slot of $\texttt{final\_out}$ and $\delta_{\texttt{final\_out}}$ via per-chunk all-to-alls; the freed slot is reused as the destination for the corresponding chunk's intermediate buffer, exactly as in the forward pass.
A new full-sized accumulator $\delta_X$ ($1$ unit) for the gradient w.r.t. the layer input is allocated at the end of stage~0's inverse all-to-all and updated by every subsequent stage.
$\delta_X$ cannot share storage with a freed slot of \texttt{final\_out} or $\delta_{\texttt{final\_out}}$: each per-stage matmul $\delta Q^i \cdot W_q^{(i)}$ contributes to \emph{all} $d_{\textit{model}}$ components of $\delta_X$ at once, so $\delta_X$ must be a contiguous full-sized buffer that exists from stage~0 onward, while slots are freed only one at a time.

The \texttt{inp\_all\_to\_all} column reports the peak at the end of stage~0, immediately after the inverse all-to-all returns and $\delta_X$ is allocated for the first time.
At that moment, the live tensors are $X$ ($1$), $\delta_X$ ($1$), the remaining slots of \texttt{final\_out} and $\delta_{\texttt{final\_out}}$ after slot~0 has been consumed (each $(\nu-1)/\nu$), and the Q-side gradient $\delta Q^0$ together with the per-group key/value accumulators $\delta K^0, \delta V^0$ ($\gamma/\nu$ combined), giving
\[
2 \;+\; \gamma/\nu \;+\; 2(\nu-1)/\nu.
\]
At $\nu = 1$ this collapses to $\gamma + 2$, matching Ulysses with activation offloading.

The FlashAttention backward kernel column reports the maximum of two per-stage formulas to capture the global backward peak across all stages.
The first argument, $3 + 2\gamma/\nu$, is the stage-0 peak: $X + \texttt{final\_out} + \delta_{\texttt{final\_out}} + (\delta Q^{\prime 0}, \delta K^{\prime 0}, \delta V^{\prime 0}) + (Q^{\prime 0}, K^{\prime 0}, V^{\prime 0})$ (no $\delta_X$ yet).
The second argument, $2 + 2\gamma/\nu + 2(\nu-1)/\nu$, is the stage-1 peak with $\delta_X$ live and slot~0 of $\texttt{final\_out}, \delta_{\texttt{final\_out}}$ already consumed: $X + \delta_X + (\nu-1)/\nu\,(\texttt{final\_out} + \delta_{\texttt{final\_out}}) + (\delta Q^{\prime 1}, \delta K^{\prime 1}, \delta V^{\prime 1}) + (Q^{\prime 1}, K^{\prime 1}, V^{\prime 1})$.
At $\nu = 1$ stage~1 does not exist; the cell equals $2\gamma + 3$ (corresponding to stage~0), matching Ulysses with offloading.
At $\nu = 2$ the two arguments are equal; for $\nu \geq 3$ the stage~1 form is strictly larger and dictates the cell value.

The \texttt{out\_all\_to\_all} column is reported w.r.t stage~0 equating to $3 + 1/\nu$, with $\nu = 1$ corresponding to Ulysses with offloading; for $\nu \geq 2$ its per-stage peak is also $1$ unit larger when $\delta_X$ is live, but this is not the global peak.
The \emph{before backward attention} column remains constant (at $3$) because $\delta_X$ has not yet been allocated.

\section{Memory usage comparison}

\label{app:b}

\begin{table*}[h]
\centering
\caption{\textbf{Memory comparison} (GiB) for \textbf{Llama 3-8B} (8$\times$H100s) and \textbf{Qwen3-32B} (16$\times$H100s) across varying sequence lengths. \textbf{OOM} denotes Out of Memory. FPDT execution fails at lengths $>$ 4M. The lowest memory usage at a given sequence length is in bold.}
\label{tab:appendix-performance}
\begin{tabular}{llcccccccccc}
\toprule
Model&Method&128K&256K&512K&1M& 2M& 3M& 4M& 5M\\
\midrule

\multirow{5}{*}[0em]{Llama 3-8B}
&Native PyTorch\phantom{x}& 25.32& 31.40& 43.55& 67.86& OOM& -- & -- & --\\
&Ring& 21.32& 23.40& 27.58& 35.86& 52.49& 69.11& OOM & --\\
&Ulysses& 21.26& 23.02& 26.80& 34.35& 49.49& 64.55 & OOM & --\\
&FPDT& 21.73& 22.50&\bf 24.03&\bf 27.09&\bf 35.17&\bf 43.35&\bf 51.42 & --\\
&\ours{}&\bf 21.10&\bf 22.30& 24.70& 29.90& 40.50& 51.10& 61.70 &\bf 72.30\\
\midrule

\multirow{5}{*}[0em]{Qwen3-32B}
&Native PyTorch\phantom{x}& 45.81& 53.69 & 69.47& OOM& -- & -- & -- & --\\
&Ring& 40.14& 41.16& 44.22& 50.51 & 63.11 & OOM & -- & --\\
&Ulysses& 40.13& 41.16& 44.10& 50.27& 62.60& OOM & -- & --\\
&FPDT&\bf 38.94 &\bf 39.47&\bf 40.54&\bf 42.66&\bf 46.91&\bf 52.27&\bf 57.77 & --\\
&\ours{}& 39.98& 40.84& 42.72 & 46.84 & 55.65 & 64.47& 73.28 & OOM\\

\bottomrule
\end{tabular}
\label{tab:app_mem}
\end{table*}

Table \ref{tab:app_mem} shows the memory usage comparison of various context parallelism schemes on Llama 3-8B and Qwen3-32B models across different context lengths. 
FPDT exhibits the best memory usage, but performs poorer due to the CPU overhead. 
UPipe has better memory efficiency than all other methods, while also matching the throughput of Ulysses. 
Note that while FPDT reports lower allocated memory, it is unable to run with context lengths greater than 4M.

\newpage
\section{Runtime comparison between DeepSpeed-Ulysses and UPipe}

\label{app:c}

\begin{table*}[h]
\centering
\caption{Runtime comparison (in seconds) for Llama 3-8B on 8$\times$H100 between DeepSpeed-Ulysses and UPipe across sequence lengths. \textbf{FA3-Fwd/Bwd:} Total Flash Attention-3 forward and backward kernel time. \textbf{Total} refers to the total time for a single training step.}
\label{tab:runtime}
\begin{tabular}{llcccccccccc}
\toprule
Method & Stage &128K&256K&512K&1M&2M&3M\\
\midrule 
\multirow{5}{*}[0em]{Ulysses}

& All-to-All&0.40& 0.90& 1.68& 4.93& 16.30& 42.21\\
& FA3-Fwd& 1.58& 6.35& 25.71& 103.49& 421.67& 995.92\\
& FA3-Bwd& 2.40& 9.13& 36.74& 146.86& 588.73& 1324.71\\
& Other& 3.03& 5.33& 10.08& 19.78& 41.30& 56.31\\
& \textbf{Total}& \textbf{7.40}& \textbf{21.72}& \textbf{74.21}& \textbf{275.06}& \textbf{1068.00}& \textbf{2419.14}\\
\midrule

\multirow{5}{*}[0em]{UPipe}

&All-to-All&0.46& 1.10& 2.43& 5.52& 17.12& 34.34\\
&FA3-Fwd& 1.51& 6.38& 25.93& 103.92& 417.55& 940.62\\
&FA3-Bwd& 2.41& 9.25& 36.99& 147.37& 590.79& 1330.76\\
&Other& 2.82& 5.23& 10.10& 19.58& 37.76& 55.52\\
&\textbf{Total}& \textbf{7.20}& \textbf{21.96}& \textbf{75.45}& \textbf{276.39}& \textbf{1063.23}& \textbf{2361.24}\\
\bottomrule
\end{tabular}
\end{table*}

Table \ref{tab:runtime} shows the runtime comparison for Llama 3-8B on a single 8$\times$H100 node between DS-Ulysses and UPipe. It shows the breakdown of the runtime of a single training step into major components: Flash Attention-3 forward time, Flash Attention-3 backward time, and All-to-All communication time. 
Note that UPipe has a higher runtime at shorter sequence lengths due to multiple kernel launches. 
At longer sequence lengths, this disadvantage is amortized, because every kernel executes more computational operations and starts to operate in the compute-bound regime.
Note that the table above reports the time per training step, so it exhibits greater variance than the throughput numbers in Table~\ref{tab:performance} (especially at shorter sequence lengths).

\section{Training Loss Convergence}

\begin{table}[h]
    
    \centering
    \caption{Loss Convergence: Comparison of training loss between UPipe and Ulysses when training a Llama 3-8B model on C4 dataset with sequence length of 128K tokens over 1000 steps.}
    \begin{tabular}{lcccccccccc}
    \toprule
    Method & 100 & 200 & 300 & 400 & 500 & 600 & 700 & 800 & 900 & 1000 \\
    \midrule
    Ulysses & 6.83 & 6.58 & 5.98 & 5.62 & 5.22 & 4.88 & 4.94 & 5.11 & 4.68 & 4.47 \\
    UPipe   & 6.85 & 6.54 & 5.98 & 5.62 & 5.22 & 4.90 & 4.95 & 5.11 & 4.69 & 4.47 \\
    \midrule
    \%Diff  & 0.29\% & -0.61\% & 0.00\% & 0.00\% & 0.00\% & 0.41\% & 0.20\% & 0.00\% & 0.21\% & 0.00\% \\
    \bottomrule
    \end{tabular}
    \label{tab:loss_convergence}
\end{table}

UPipe performs the exact same computation during the attention stage as Ulysses. However, our proposed GQA scheduling strategy changes the order in which different attention heads are processed. This could lead to minor numerical differences due to the non-associative nature of floating point operations. 
In this section, we verify that UPipe does not affect the training loss convergence by comparing the training dynamics with DeepSpeed-Ulysses on Llama 3-8B. 

We train this model on a single 8$\times$H100 node with a sequence length of 128K tokens, a batch size of 1, and a learning rate of $10^{-5}$. We use the C4 dataset~\cite{c4} for training the model on 1000 steps, and report the loss after every 100 steps in Table \ref{tab:loss_convergence}.

\section{Scope and Limitations}
\label{app:limitations}

\ours{} applies broadly to Transformer-based models (LLMs, Vision Transformers, Diffusion Transformers) regardless of the modality, as the only component it modifies is multi-head self-attention. 
Although its throughput at shorter contexts is lower than DeepSpeed-Ulysses, better memory efficiency enables larger batch sizes and Selective Activation Checkpointing (Appendix \ref{app:sac}). 
Using an all-to-all implementation that does not involve Streaming Multiprocessors (also known as ``SM-free communication'') could further improve throughput by overlapping communication with attention computation.

\end{document}